\documentclass[sn-mathphys-num]{sn-jnl}
\usepackage{graphicx}
\usepackage{multirow}%
\usepackage{amsmath,amssymb,amsfonts}%
\usepackage{amsthm}%
\usepackage{mathrsfs}%
\usepackage[title]{appendix}%
\usepackage{xcolor}%
\usepackage{textcomp}%
\usepackage{manyfoot}%
\usepackage{booktabs}%
\usepackage{algorithm}%
\usepackage{algorithmicx}%
\usepackage{algpseudocode}%
\usepackage{listings}%
\usepackage{caption}%

\begin{document}

\title[Article Title]{Robust Counterfactual Explanations under Model Multiplicity Using Multi-Objective Optimization}

\author{\fnm{Keita} \sur{Kinjo}}\email{kkinjo@kyoritsu-wu.ac.jp}

\affil{\orgdiv{Faculty of Business}, \orgname{Kyoritsu Women's University}, \orgaddress{\street{2-2-1 Hitotsubashi}, \city{Chiyoda-ku}, \postcode{101-8437}, \state{Tokyo}, \country{Japan}}}


\abstract{In recent years, explainability in machine learning has gained importance. In this context, counterfactual explanation (CE), {which presents the minimal change to input features required to alter a model's prediction to a desired target,} has attracted attention. However, it has been pointed out that CE is not robust when there are multiple machine-learning models with similar accuracy {(model multiplicity)}. {Since users should not be penalized by an arbitrary model choice, generating CEs that remain stable across models is essential for reliable decision-making under model multiplicity.} In this paper, we propose robust CEs that introduce a new viewpoint---Pareto improvement---and a method that uses multi-objective optimization to generate {them}. To evaluate the proposed method, we conducted experiments using both simulated and real data. {Real-data experiments on two domains---web search trend data with continuous features and educational intervention data with binary features---demonstrate that the proposed method produces robust CEs under model multiplicity and is applicable to diverse real-world settings.} This study highlights the potential of ensuring robustness in decision-making by applying the concept of social welfare. We believe that this research can serve as a valuable foundation for various fields, including explainability in machine learning, decision-making, and action planning based on machine learning.}


\maketitle
\noindent\small
\textbf{Conflicts of Interest:} The authors declare no conflicts of interest.
\smallskip

\noindent\small
\textbf{Acknowledgements:} This study was supported by JSPS KAKENHI Grant-in-Aid for Scientific Research (C) JP25K05381.

\newpage

\section{Introduction}\label{sec1}

{Artificial intelligence (AI), including machine learning, is used in many domains. However, although many machine-learning methods have high prediction accuracy, they are often considered ``black boxes'' because the processes involved are unclear owing to their complex combination of nonlinearities and interactions. Explainable AI or interpretable machine learning has become an important \textit{research area} in addressing these problems \cite{linardatos2020, angelov2021, dwivedi2023}. Several such methods are available. White-box models (e.g., linear regression, decision trees) expose their internal parameters and are directly interpretable. In contrast, black-box models (e.g., deep neural networks, ensemble methods) achieve high predictive accuracy but do not reveal how predictions are made, motivating the need for post-hoc explanation methods, among others. Among these, methods that examine which variables are globally important across the entire dataset (global explanations) and those that explain predictions for individual instances (local explanations) have been proposed. One prominent local explanation method is the counterfactual explanation (CE) \cite{karimi2022, guidotti2024, verma2024}.}

{CEs are outputs that indicate, for a trained supervised machine-learning model, the minimal changes to the original input features needed to achieve a particular desired predictive outcome---a concept known as \textit{proximity}. This clarifies the factors influencing the forecast and improves the explainability of the model. For example, if a person is denied a loan, a CE will suggest how to change specific features (e.g., increase annual income by a certain amount) to be approved; the CE does not assign blame to a single feature (which is the domain of feature-attribution methods such as LIME \cite{linardatos2020}), but rather recommends a combination of feature changes needed to achieve a desired outcome. This method is important because it can provide suggestions for machine learning users regarding the actions they should take. CEs are also called \textit{algorithmic recourse} \cite{karimi2022}. It has also been noted that CEs are related to \textit{adversarial examples} \cite{freiesleben2022, pawelczyk2022}.}

When extracting CEs, the basic condition is to make the original data and generated CEs as similar as possible in terms of {\textit{proximity} (i.e., minimizing a distance measure between the original instance and the CE)}. In addition, various other conditions have been proposed for CEs, such as closeness to the training data/plausibility, actionability (feasibility), sparsity, diversity, and so on \cite{guidotti2024}. In addition, several extraction methods have been developed depending on the availability of access to the model and the assumptions of the model's functions (linearity, differentiability, etc.) \cite{verma2024}.

However, the robustness of CEs has long been problematic \cite{verma2024}. Robustness can be considered in various ways. Jiang et al. classified the robustness of CEs into four categories \cite{jiang2024}.

(i) \textbf{Robustness against model changes} \cite{upadhyay2021, pawelczyk2022, krishna2023}: {Whenever the model $M$ changes to $M'$ and this change is sufficiently small (e.g., due to retraining on similar data), the CE $x'$ remains valid under the updated model, i.e., $M(x') = M'(x')$ \cite{jiang2024}. For example, if a loan applicant implements the recommended income increase and the bank later retrains its model on new data, the CE should still result in loan approval.}

(ii) \textbf{Robustness against model multiplicity} \cite{pawelczyk2020}: {When multiple models achieve similar predictive accuracy on the same dataset, users should not be penalized by which model an organization happens to deploy. This definition requires that CEs remain consistent across such models. If a user receives different recourse depending on the deployed model, their ability to act is undermined. For example, consider a loan applicant whose profile is fixed; under one equally accurate model the CE requires an income increase of \pounds5{,}000, whereas under another equally accurate model the same applicant would need \pounds15{,}000---the discrepancy arising solely from which model the organisation happens to deploy.}

(iii) \textbf{Robustness against noisy executions} \cite{pawelczyk2023}: Robust CEs are extracted such that, even if their attributes change slightly {during execution (e.g., due to stochastic decision processes)}, the predictions do not change significantly. {For example, if a CE recommends increasing annual income by \pounds5{,}000 but the actual increase is \pounds4{,}800 due to rounding, the loan application should still be approved.}

(iv) \textbf{Robustness against input changes} \cite{slack2021}: Robust CEs are extracted under the condition that if the predictions of two similar data points are identical, then the CEs of the data are also similar {(i.e., stability under small perturbations of the input). For example, two loan applicants with nearly identical financial profiles who both receive the same rejection should be offered similar CEs.}

Regarding (ii), the existence of multiple models with similar accuracy, which is the premise of the problem, is an important issue {that arises in practice when multiple learners are trained on the same dataset \cite{breiman2001}. This phenomenon, sometimes called the Rashomon effect \cite{breiman2001}, is commonly observed when} multiple learners of different classes are trained on the same dataset. Furthermore, in such situations, CE is an important task for social applications because it can be used to identify important variables without selecting a model and to ensure safety when making further decisions. For example, it is important to select the most effective CEs when making decisions based on CEs for {safety-critical} issues such as medical care or issues involving huge costs such as marketing. However, the number of relevant studies is limited \cite{pawelczyk2020, leofante2023, jiang2024}.

Pawelczyk et al.\ \cite{pawelczyk2020} first analyzed CE robustness under model multiplicity, demonstrating a theoretical cost--robustness trade-off between sparse and data-support methods. Leofante et al.\ \cite{leofante2023} proposed exact methods for finding CEs valid across all models, but restricted to white-box neural network classifiers. Jiang et al.\ \cite{jiang2024recourse} proposed an argumentation-based ensembling approach, limited to classification tasks. These studies leave open the problem of generating robust CEs for black-box models with continuous targets and flexible constraints (see Section~\ref{sec_related} for a detailed review).

To address this gap, we draw on the concept of Pareto improvement from welfare economics and multi-objective optimization theory. A Pareto improvement is a change that benefits at least one {objective} without worsening any other~\cite{pareto2014}. The Pareto front is the set of solutions from which no further Pareto improvement is possible, providing decision-makers with a diverse set of robust options. As will be discussed in more detail below, this idea motivates a CE that achieves consistent improvement across all models. This idea is important when choosing {robust and actionable} solutions based on the CE, which is a costly and {safety-critical} problem in society.

In this paper, we propose robust CEs for model multiplicity by introducing a new viewpoint, Pareto improvement, and multi-objective optimization. In addition, we devised a validation index and verified its robustness through simulated data and its {applicability} through two real-world datasets covering both continuous features (web search trend data) and binary features (educational intervention data). {The proposed method has four key features. First, it employs the concept of Pareto improvement to address the problem of model multiplicity, providing a theoretically grounded approach to robust CE generation. Second, unlike prior work, it can be applied not only to classification but also to continuous target variables (regression), and supports flexible inclusion of constraint conditions. Third, by extracting a diverse set of CEs on the Pareto front, it allows users to select solutions according to their preferences or constraints. Fourth, the method is validated on two real-world datasets---web search trend data with continuous features and educational intervention data with binary features---demonstrating broad applicability across diverse problem settings.}

Detailed comparisons with prior work are provided in Section~\ref{sec_related}. By grounding the approach in Pareto improvement and multi-objective optimization, this research contributes to explainability in machine learning, decision-making, and action planning.

Section {2} presents the proposed method. Section {3} presents the validation using simulated and real data. Section {4} provides a discussion. Section {5} describes limitations. Section {6} reviews related work, and Section {7} provides conclusions.

\section{Method}\label{sec2}

In the following sections, Section {2.1} describes the problem setup, Section {2.2} provides the multi-objective optimization framework that underpins the proposed method, Section {2.3} describes the proposed method, and Section {2.4} explains the evaluation method.

\subsection{Problem Setting}\label{subsec1}

We set up our problem as follows. We have data $\mathcal{D}$, consisting of $n$ pairs of $y_i$ (scalars) and $X_i$ ($r$-dimensional vectors), where $i$ is an index per sample. Let $y_i \in \mathcal{Y} \subseteq \mathbb{R}$ and $X_i \in \mathcal{X} \subseteq \mathbb{R}^r$. $\mathcal{Y}$ and $\mathcal{X}$ are feature spaces. {Here $X_b \in \mathcal{X}$ denotes the base (input) instance to be explained, and $y_t \in \mathcal{Y}$ denotes the desired target value.}

\begin{equation}\label{eq:dataset}
\mathcal{D} = \left\{(y_i, X_i)\right\}_{i=1}^{n}
\end{equation}

There are $m$ machine-learning models $f_{j=1, \dots, m}: \mathcal{X} \to \mathcal{Y}$ estimated from $\mathcal{D}$, {selected as the top $m$ models ranked by prediction accuracy (see Section~\ref{subsec3})}. Based on the above, the objective of this study is to obtain the solution $X_{cf}^*$ to the following problem:

\begin{equation}\label{eq:main_obj}
X_{cf}^* = \operatorname*{argmin}_{X_{cf} \in \mathcal{X}} \left( \text{loss}(y_t, f_1(X_{cf})), \text{loss}(y_t, f_2(X_{cf})), \dots, \text{loss}(y_t, f_m(X_{cf})) \right)
\end{equation}

subject to

\begin{equation}\label{eq:prox}
d(X_b, X_{cf}) \leq C,
\end{equation}

\begin{equation}\label{eq:ineq}
g_j(X_{cf}) \geq 0, \quad j = 1, \dots, J,
\end{equation}

\begin{equation}\label{eq:eq}
h_k(X_{cf}) = 0, \quad k = 1, \dots, K.
\end{equation}

where $y_t \in \mathcal{Y}$ is a target value, $X_b \in \mathcal{X}$ is a base data to be explained, and $X_{cf}$ is a candidate for counterfactual explanations. $\text{loss}$ is the loss function between $y_t$ and $f_j(X_{cf})$. Specifically, when $y$ is a continuous variable, the squared or absolute error is used. When $y$ is a categorical variable, the cross-entropy error is used. The $m$ values of the losses are vectors. $d$ is a distance function {(e.g., Euclidean distance $\|X_b - X_{cf}\|_2$)}. $C$ is the upper bound of the distance function, {serving as the \textit{proximity constraint} that ensures the CE does not deviate excessively from the original instance.} {Typical inequality constraints $g_j$ include lower- and upper-bound constraints on individual features (e.g., $g_j(X_{cf}) = X_{cf}[j] - \mathrm{lb}_j \geq 0$) or monotonicity constraints. Equality constraints $h_k$ can fix certain features that are not actionable (e.g., $h_k(X_{cf}) = X_{cf}[k] - X_b[k] = 0$).} In summary, this study seeks to find $X_{cf}$ such that the {distance $d(X_b, X_{cf})$ is less than or equal to $C$}, satisfies certain constraints, and is better than the other solutions in the loss function for all models. Such a solution with repeated Pareto improvements is called the Pareto {front}; the formal definition is given in Section~\ref{subsec2}.

\subsection{Multi-Objective Optimization}\label{subsec2}

\sloppy

This section provides the theoretical foundation for the proposed method, introducing the Pareto front and the NSGA-II algorithm \cite{deb2016}. We introduce $L$ ($l = 1, \dots, L$) objective functions $F_l: E \to \mathbb{R}$ corresponding to $r$-dimensional variables $\theta \in E$. Let $E$ be a domain and $E \subseteq \mathbb{R}^r$. Let $F: E \to \mathbb{R}^L$ be the function that summarizes them. We solve the following optimization problem:

\begin{equation}\label{eq:moo}
\min_{\theta \in C} F(\theta), \quad \text{where} \quad F(\theta) = \left(F_1(\theta), F_2(\theta), \dots, F_L(\theta)\right).
\end{equation}

Constraints can also be included. The Pareto solution, which is the solution to this problem, refers to a $\theta^*$ such that there exists no $\theta$ for which $F(\theta) \leq F(\theta^*)$ and $F(\theta) \neq F(\theta^*)$. This is also called a non-dominated or efficient solution. The condition $F(\theta) \leq F(\theta^*)$ means that $F_l(\theta) \leq F_l(\theta^*)$ for all $l$. A move from a dominated solution to a Pareto solution constitutes a Pareto improvement. {The set of all such solutions forms the \textit{Pareto front} $\Theta^* \ni \theta^*$ in the objective function space. In multi-objective optimization, one obtains a set of non-dominated solutions (the Pareto front) rather than a single solution. This diversity is valuable, as it allows decision-makers to select among multiple trade-off options.}

There are many methods for computing Pareto solutions in multi-objective optimization (MOO). For example, there are methods that use a weighted sum of multiple objective functions and the $\epsilon$-constraint method, which optimizes a specific objective function and defines all other functions as constraints. In addition to those that seek a specific optimal solution, there are also those that seek a set of Pareto solutions, as described above \cite{deb2011}.

Recently, several methods have been proposed for computing Pareto solution sets, including evolutionary computation and descent methods. A well-known method that uses evolutionary computation is the fast elitist non-dominated sorting genetic algorithm (NSGA-II) \cite{fliege2000, deb2002}. This method identifies a set of solutions by iteratively performing ranking, selection, crossover, and mutation operations on a population of candidate solutions. During this process, the crowding distance metric is utilized to ensure a diverse spread of solutions across the Pareto front.

Because evolutionary computation is an approximate solution method (heuristic), there is no guarantee that a Pareto solution set can be obtained. Therefore, the output is often a non-dominated, non-inferior solution set among the solution sets obtained in the solution search process. However, methods using evolutionary computation have the advantage of being able to extract a wide variety of solutions, are robust to nonlinear and nonconvex functions, and eliminate the need to differentiate the objective function. However, this approach is computationally expensive. A comparison with other MOO approaches is provided in Section~\ref{sec_related}.

\subsection{Proposed Method}\label{subsec3}

We describe the procedures proposed in this paper. {Building on the MOO framework and evolutionary computation introduced in Section~\ref{subsec2}, we outline Processes~1--4 below (summarised in Algorithm~\ref{alg:moo_ce}). Concrete data-split ratios, model hyperparameters, and experiment-specific settings are reported in Section~\ref{sec3}.}

1. Split $\mathcal{D}$ into training data $\mathcal{D}_{\text{train}} \subset \mathcal{D}$ and test data $\mathcal{D}_{\text{test}} \subset \mathcal{D}$ ($\mathcal{D}_{\text{train}} \cap \mathcal{D}_{\text{test}} = \emptyset$).

2. Set up $M$ models in advance. Estimate each model $f_j$ based on $\mathcal{D}_{\text{train}}$ and calculate the accuracy of prediction using $\mathcal{D}_{\text{test}}$.

3. Select $m$ models based on their accuracy.

4. Derive $S$ solutions $X_{cf,s}^*$ ($s = 1, \dots, S$) for $m$ models by multi-objective optimization.

In Process 2, $M$ various models are prepared in advance. The {mean squared error (MSE; lower values indicate better fit)} is used {to evaluate prediction quality} because continuous variables are used for $y$ in this study. In Process 3, sorting is performed based on {MSE}, and the top $m$ models are used. {By selecting the top-$m$ models by accuracy, we avoid including poorly-fitting models that would introduce noise into the multi-objective optimization. In particular, models with similar MSE values are more likely to define a coherent Pareto front, since their prediction surfaces overlap in the regions of interest; conversely, including a model with substantially higher MSE (lower quality) may skew the Pareto front towards directions that do not generalise under the true function. An ablation study varying $m \in \{2, 3, 4\}$ is reported in Experiment~1 (Section~\ref{subsec5}).} Another possible method is to set a threshold value and select models that exceed it. This selection may be a response to the increase in dissimilarity caused by the introduction of a Pareto solution. For Processes 1--3, it is possible to select the top $m$ models from a large number of models using automated machine learning. In Process 4, NSGA-II is used because the machine-learning function is nonlinear, model-independent, and derives a wide variety of solutions. After Process 4, it is possible to select a solution among the $S$ solutions according to the user's preference or to select the solution closest to all solutions (medoid, close to centroid, etc.) to select a safer solution.

{
\begin{algorithm}[t]
\caption{MOO-CE: Robust CE Generation under Model Multiplicity}\label{alg:moo_ce}
\begin{algorithmic}[1]
\Require Dataset $\mathcal{D}$, base instance $X_b$, number of models $m$, proximity bound $C$, NSGA-II hyperparameters
\Ensure Pareto front of robust CEs $\{X_{cf,s}^*\}_{s=1}^{S}$
\State Split $\mathcal{D}$ into $\mathcal{D}_{\text{train}}$ and $\mathcal{D}_{\text{test}}$
\State Train $M$ models $\{f_j\}_{j=1}^{M}$ on $\mathcal{D}_{\text{train}}$; evaluate MSE on $\mathcal{D}_{\text{test}}$
\State Select top-$m$ models $\{f_1, \dots, f_m\}$ ranked by MSE (ascending)
\State Define multi-objective problem:
\Statex \quad $\min_{\Delta X} \bigl(\text{loss}(y_t, f_1(X_b + \Delta X)),\; \dots,\; \text{loss}(y_t, f_m(X_b + \Delta X))\bigr)$
\Statex \quad s.t.\ $\|\Delta X\|_2 \leq C$,\; $g_j(X_b + \Delta X) \geq 0$,\; $h_k(X_b + \Delta X) = 0$
\State Run NSGA-II to obtain Pareto front $\mathcal{P} = \{\Delta X_s^*\}$
\For{each $\Delta X_s^* \in \mathcal{P}$}
    \State $X_{cf,s}^* \gets X_b + \Delta X_s^*$
\EndFor
\State \Return $\{X_{cf,s}^*\}_{s=1}^{S}$ \Comment{User selects preferred CE from Pareto front}
\end{algorithmic}
\end{algorithm}
}

To theoretically demonstrate the robustness of the proposed method, we compare it with a simple approach based on the weighted sum of loss functions. In weighted sum optimization, only a single solution can be obtained, and depending on the choice of weights, one of the loss functions may deteriorate. For this discussion, we assume two models with identical accuracy and that $X_{cf}$ is one-dimensional.

We consider minimizing a linear combination of the loss functions using weights $w_1, w_2 (>0)$. For simplicity, we assume there are no constraints.

\begin{equation}\label{eq:wsum}
X_{cf}^\ast = \underset{X_{cf} \in \mathcal{X}}{\arg\min} \left( w_1 \text{loss}(y_t, f_1(X_{cf})) + w_2 \text{loss}(y_t, f_2(X_{cf})) \right).
\end{equation}

At the optimal solution $X_{cf}^\ast$, the first-order necessary condition gives:

\begin{equation*}
\frac{d}{dX_{cf}} \left( w_1 \text{loss}(y_t, f_1(X_{cf}^\ast)) + w_2 \text{loss}(y_t, f_2(X_{cf}^\ast)) \right) = 0
\end{equation*}

\begin{equation*}
w_1 \frac{d}{dX_{cf}} \text{loss}(y_t, f_1(X_{cf}^\ast)) + w_2 \frac{d}{dX_{cf}} \text{loss}(y_t, f_2(X_{cf}^\ast)) = 0
\end{equation*}

\begin{equation*}
\frac{d}{dX_{cf}} \text{loss}(y_t, f_1(X_{cf}^\ast)) = -\frac{w_2}{w_1} \frac{d}{dX_{cf}} \text{loss}(y_t, f_2(X_{cf}^\ast))
\end{equation*}

Thus, $X_{cf}^\ast$ is determined by the weight ratio $w_1 / w_2$, and near the optimal value, the gradients of the respective loss functions have opposite signs.

Next, we consider the case where the loss functions are either monotonically increasing or monotonically decreasing in the neighborhood of $X_{cf}^\ast$. That is, for some $X_b < X_{cf}^\ast$,

\begin{equation*}
\frac{d}{dX_{cf}} \text{loss}(y_t, f_1(X_{cf})) < 0, \quad \frac{d}{dX_{cf}} \text{loss}(y_t, f_2(X_{cf})) > 0
\end{equation*}

holds. Under this monotonicity assumption, we obtain:

\begin{equation*}
\text{loss}(y_t, f_1(X_{cf}^\ast)) \leq \text{loss}(y_t, f_1(X_b)), \quad
\text{loss}(y_t, f_2(X_{cf}^\ast)) \geq \text{loss}(y_t, f_2(X_b)).
\end{equation*}

Similarly, when the monotonicity of $\text{loss}(y_t, f_1(X_{cf}))$ and $\text{loss}(y_t, f_2(X_{cf}))$ are reversed, the same relationship holds.
This implies that, under the stated monotonicity assumption, improving one loss function may lead to the deterioration of the other---a common failure mode of weighted-sum scalarization when the loss surfaces have opposing gradients near the optimum.

On the other hand, in multi-objective optimization, optimal solutions are sought while considering all loss functions simultaneously, ensuring that no function deteriorates by definition. Additionally, multiple trade-off solutions (the Pareto front) can be obtained, allowing for the selection of an appropriate solution.

\subsection{Evaluation Method}\label{subsec4}

In this study, we developed evaluation indices drawing on established CE metrics from the literature \cite{verma2024, guidotti2024}---such as validity, proximity, and plausibility---and extended them with a novel robustness index (TIR) to assess whether CEs lead to genuine improvements under the true function.

\textbf{Validity (\textit{Val})}: This is an evaluation of the closeness of the prediction by CE to the desired value \cite{verma2024}. {Specifically, it measures whether the CE changes the model output towards the target $y_t$.} In the general case with a finite target $y_t$, Val is defined as the mean loss:
\begin{equation}\label{eq:val}
    \textit{Val}_j = \frac{\sum_{s=1}^{S} |y_t - f_j(X_{cf,s}^*)|}{S}
\end{equation}
{In this study, we set $y_t = \infty$ because the goal is to maximize the predicted value rather than reach a specific target. In this case, Val reduces to the mean predicted value $\frac{1}{S}\sum_{s=1}^{S} f_j(X_{cf,s}^*)$, where a higher value indicates a better CE.}

{\textbf{Val Improvement ($\Delta$\textit{Val})}: The mean predicted value of the CE minus the predicted value of the base instance. This is essentially \textit{Val} expressed relative to the base, making it easier to interpret across datasets with different scales. A larger value indicates a greater predicted benefit from applying the CE. This indicator is primarily used in the real-data experiments (Experiment~3), where the true function is unknown and TIR cannot be computed.}
\begin{equation}\label{eq:dval}
    \Delta\textit{Val}_j = \frac{\sum_{s=1}^{S} f_j(X_{cf,s}^*)}{S} - f_j(X_b)
\end{equation}

\textbf{Dissimilarity (\textit{Dissim})}: This indicator is the opposite of proximity. {It measures the distance between the CE and the original instance; a smaller value corresponds to a lower cost of change.} The smaller this indicator, the better the CE because it corresponds to the cost.
\begin{equation}\label{eq:dissim}
    \textit{Dissim} = \frac{\sum_{s=1}^{S} \|X_{cf,s}^* - X_b\|_2}{S}
\end{equation}

\textbf{Plausibility (\textit{Plaus})}: This indicator {measures the closeness of the CE to the training data distribution (nearest-neighbour distance), assessing how realistic the CE is. We avoid using the term ``feasibility,'' which typically refers to whether a change is practically achievable by a user, to prevent confusion.} The index is based on \cite{guidotti2024} and others. The smaller this indicator is, the more {plausible} it is, which makes it a good CE. It is used in constraints and as an evaluation indicator \cite{verma2024}.
\begin{equation}\label{eq:plaus}
    \textit{Plaus} = \frac{\sum_{s=1}^{S} \min_{X_i \in X} \|X_{cf,s}^* - X_i\|_2^2}{S}
\end{equation}

\textbf{True Improvement Ratio (\textit{TIR})}: We propose the following method as an indicator to check robustness: {Let $y_s = y_\mathrm{true}(X_{cf,s}^*)$ denote the true outcome at CE $s$, and $y_0 = y_\mathrm{true}(X_b)$ denote the true outcome at the base instance. TIR measures the proportion of CEs for which the true function value genuinely exceeds the base value.} If this value is high, then the CE is robust. {Note that this indicator can only be used when the true function is known, limiting its application to simulated data.}
\begin{equation}\label{eq:tir}
    \textit{TIR} = \frac{\sum_{s=1}^{S} \mathbb{1}\{y_s - y_0 > 0\}}{S}
\end{equation}

When there were $S$ counterfactual explanations (CEs) for a base dataset, we used the average value described below to evaluate the method. For Method~1, $S$ equals the number of restarts (different initial values); for Method~3, $S$ equals the size of the Pareto front, which may vary across base instances. When evaluating the method as a whole, we compared the mean values of the CEs for multiple base datasets.

\section{Experiment}\label{sec3}

In Experiment 1, the robustness of the CE extracted by the proposed method was verified using simulated data; in Experiment 2, the proposed method was compared against established CE baselines to contextualise its performance; in Experiment 3, it was applied to real data and its applicability was discussed.

\subsection{Experiment 1: Simulation Data}\label{subsec5}

Here, we applied this method and other methods to the simulation data for which the true function is known and compared them. A comparison was performed on data with complex nonlinear functions.

The four models used are:

\textbf{Model 1}: Linear regression

\textbf{Model 2}: Random Forest regression with 100 trees

\textbf{Model 3}: LightGBM regression with 100 boosting rounds \cite{ke2017}

\textbf{Model 4}: Multilayer perceptron regression (MLP) with one layer of 100 units and ReLU activation

which are commonly used in machine learning. {These four models were selected for the following reasons. First, they span a broad range of model classes---linear (Model~1), tree-based ensemble (Model~2), gradient boosting (Model~3), and neural network (Model~4)---covering the main architectural families deployed in practice. Second, their inductive biases differ substantially: Linear Regression assumes a parametric linear relationship; Random Forest averages many decision trees to capture nonlinear patterns non-parametrically; LightGBM sequentially minimizes residuals via gradient boosting and achieves state-of-the-art performance on tabular data \cite{ke2017}; and MLP is a universal function approximator capable of modeling highly complex dependencies. Third, all four are widely used in real-world applications across diverse domains.}

The following three methods of extracting CEs are compared:
\begin{itemize}

    \item \textbf{Method 1}: Multiple CEs are generated by changing the initial value of each model. Specifically, the following optimization problem is solved using Sequential Least Squares Programming (SLSQP) to generate CEs:
    \begin{equation}\label{eq:method1}
    X_{cf}^{**} = \operatorname*{argmin}_{X_{cf} \in X} \left( \text{loss}(y_t, f_1(X_{cf})) + \lambda d(X_b, X_{cf}) \right),
    \end{equation}
    \[
    \text{subject to} \quad d(X_b, X_{cf}) \leq C, \quad g_j(X_{cf}) \geq 0, \quad j = 1, \dots, J,
    \]
    \[
    h_k(X_{cf}) = 0, \quad k = 1, \dots, K.
    \]
    where $\lambda$ is the {regularization coefficient controlling the trade-off between validity and proximity.} {Here $X_{cf}^{**}$ denotes the CE obtained from this single-model optimization, to distinguish it from the Pareto-front solutions $X_{cf,s}^*$ of Method 3.}
    \item \textbf{Method 2}: Multiple CEs are generated by changing the initial values based on a stacking model using the predictions of the above models. {Specifically, a stacking model is constructed by training a linear regression meta-learner on the out-of-fold predictions of Models 1--4 on $\mathcal{D}_{\text{train}}$. The meta-learner takes the four model predictions as inputs and produces a single aggregated prediction. The CE generation procedure is then the same as Method 1, applied to this stacking model.} This method can also be interpreted as a solution approach for multi-objective optimization using weighted linear summation of objective functions, provided that all coefficients are positive.
    \item \textbf{Method 3}: Multiple CEs are generated using the proposed multi-objective optimization-based method. Specifically, we compare cases in which the number of models is set to 2, 3, and 4, in descending order of {MSE-based} accuracy. For the multi-objective optimization algorithm, we adopt NSGA-II \cite{blank2020}, which is a type of evolutionary computation described in Section~\ref{subsec2}. NSGA-II is suitable for this study because it can be applied to complex functions and enables diverse solutions to be obtained by utilizing the crowding distance. {The NSGA-II hyperparameters are as follows: population size = 50, number of generations = 30 (Experiments~1 and~2) / 20 (Experiment~3), crossover probability = 0.9, mutation probability = 1/$r$ (where $r$ is the number of features), SBX crossover with distribution index $\eta_c = 20$, and polynomial mutation with distribution index $\eta_m = 20$. Solutions are represented as real-valued vectors in $\mathcal{X}$. For binary features (such as the educational interventions in Experiment~3), the optimization is performed over continuous relaxations $[0,1]$; the optimized values are retained as-is and interpreted as intervention intensities (see Section~\ref{subsec6b}).}
\end{itemize}

The model incorporates interactions and nonlinear functions as follows:

{\noindent\textbf{(1) Model with linear-dominant structure and smooth periodicity}}
{
\begin{equation}\label{eq:case1}
    y_i = \frac{x_{i,1}}{5} - \frac{x_{i,2}}{5} + \sin\!\left(\frac{x_{i,3}}{3}\right) + 0.3\cos\!\left(\frac{x_{i,4}}{3}\right) + \frac{x_{i,5}}{10} + \epsilon_i
\end{equation}
}

{\noindent\textbf{(2) Model with nonlinearity and discontinuity}}
{
\begin{equation}\label{eq:case2}
    y_i = \sin\!\left(\frac{x_{i,1}x_{i,2}}{5}\right) + \mathrm{sign}(x_{i,3})\sin(x_{i,4}) + \frac{x_{i,5}^2}{15} + \epsilon_i
\end{equation}
}

{where $x_{i,1}, x_{i,2}, x_{i,3}, x_{i,4}, x_{i,5}$ are uniform random numbers in the range $[-10, 10]$, and $\epsilon_i$ is a standard normal random variable with a mean of 0 and a variance of 1. The index $i = 1, \dots, 1000$ represents a sample. The mean and standard deviation for (1) are $-0.07$ and $2.19$, respectively, whereas those for (2) are $2.24$ and $2.40$, respectively.}

First, the accuracies of the models were compared (Table~\ref{tab:accuracy_sim}; {values to the right of $\pm$ represent the standard deviation (SD) over 20 repeated splits}). The dataset was randomly split 20 times with a training size of 0.7 and a test size of 0.3 ({data generation and model training: seed~42}). {CE generation (base-case selection and SLSQP initial values for Methods~1 and~2) used seed~0 for Case~1 and seed~99 for Case~2; NSGA-II (Method~3) used seed~1.} The mean MSEs obtained from these splits are compared below.

\begin{table}[htbp]
    \centering
    \caption{Model accuracy for Simulation Cases~1 and~2 (MSE mean $\pm$ SD over 20 splits).}
    \begin{tabular}{lcc}
        \hline
        \textbf{Model} & \textbf{Case~1 MSE (mean $\pm$ SD)} & \textbf{Case~2 MSE (mean $\pm$ SD)} \\
        \hline
        Model 1 (Linear Regression) & $1.291 \pm 0.077$ & $5.810 \pm 0.246$ \\
        Model 2 (Random Forest)     & $1.299 \pm 0.081$ & $2.080 \pm 0.142$ \\
        Model 3 (LightGBM)          & $1.263 \pm 0.065$ & $1.977 \pm 0.122$ \\
        Model 4 (MLP)               & $1.263 \pm 0.086$ & $2.155 \pm 0.150$ \\
        Stacking Model              & $1.397 \pm 0.087$ & $2.189 \pm 0.156$ \\
        \hline
    \end{tabular}
    \label{tab:accuracy_sim}
\end{table}

{In Case 1 (Table~\ref{tab:accuracy_sim}), the MSE values of all five models range from 1.263 to 1.397, with fully overlapping standard deviation ranges (mean $\pm$ SD), constituting a \textbf{valid model-multiplicity setting} in which all models---including LR---are approximately equally accurate. In Case 2 (Table~\ref{tab:accuracy_sim}), the Case~2 true function is designed so that the three nonlinear models (RF: 2.080, LGB: 1.977, MLP: 2.155) have overlapping mean $\pm$ SD ranges---constituting a \textbf{valid model-multiplicity setting among the nonlinear models}---while LR (5.810) is substantially worse and is excluded by the top-$m$ selection in Method~3. Together, Case~1 tests full model multiplicity (all models similar) and Case~2 tests nonlinear-only multiplicity (LR excluded).}

Next, we evaluated these methods (Table~\ref{tab:ta2n1}, Table~\ref{tab:ta2n2}). Specifically, we set $y_t = \infty$ {(i.e., the CE is optimized to maximize the predicted outcome; the loss function is the absolute error, $\text{loss}(y_t, f(X)) = |y_t - f(X)|$, so that minimizing the loss with $y_t = \infty$ is equivalent to maximizing $f(X)$)}. 10 CEs ($=S$) were generated for each randomly selected base. Using these multiple CEs, the evaluation metrics described in Section~\ref{subsec4} were calculated. Finally, the mean values of {20} cases were compared.

\begin{table*}[htbp]
    \centering
    \caption{Evaluation metrics for CEs --- Simulation Case~1 ({mean over 20 base instances}, $S=10$ CEs each). {Bold indicates the top-3 values per metric (all tied values included): highest for \textit{Val}, TIR, \textit{AV/D}, and \textit{AV/P}; lowest for \textit{Dissim} and \textit{Plaus}. \textit{AV/D} $=$ \textit{Ave val}/\textit{Dissim}; \textit{AV/P} $=$ \textit{Ave val}/\textit{Plaus}.}}
    \label{tab:ta2n1}
    \begin{tabular}{llrrrrrr}
        \toprule
        \textbf{Method} & & \textit{Val} & \textit{Dissim} & \textit{Plaus} & TIR & \textit{AV/D} & \textit{AV/P} \\
        \midrule
        M1 (LR)    & & 1.783 & 3.010 &  8.481 & \textbf{1.000} & 0.592 & 0.210 \\
        M1 (RF)    & &  1.252 & \textbf{2.426} & \textbf{6.647} & 0.705 & 0.516 & 0.188 \\
        M1 (LGB)   & &  1.666 & \textbf{2.193} & \textbf{7.441} & 0.765 & \textbf{0.760} & 0.224 \\
        M1 (MLP)   & &  \textbf{1.997} & 3.002 & 8.823 & \textbf{1.000} & \textbf{0.665} & \textbf{0.226} \\
        M2 (Stack) & &  0.586 & 4.821 & 122.497 & 0.020 & 0.121 & 0.005 \\
        \multirow{2}{*}{M3 ($m$=2)} & (1) & 1.614 & \multirow{2}{*}{2.918} & \multirow{2}{*}{7.866} & \multirow{2}{*}{\textbf{1.000}} & \multirow{2}{*}{0.600} & \multirow{2}{*}{0.222} \\
         & (2) & \textbf{1.886} & & & & & \\
        \multirow{3}{*}{M3 ($m$=3)} & (1) & 1.602 & \multirow{3}{*}{2.904} & \multirow{3}{*}{7.799} & \multirow{3}{*}{\textbf{1.000}} & \multirow{3}{*}{0.603} & \multirow{3}{*}{\textbf{0.225}} \\
         & (2) & 1.836 & & & & & \\
         & (3) & 1.818 & & & & & \\
        \multirow{4}{*}{M3 ($m$=4)} & (1) & 1.526 & \multirow{4}{*}{\textbf{2.843}} & \multirow{4}{*}{\textbf{7.374}} & \multirow{4}{*}{0.999} & \multirow{4}{*}{\textbf{0.627}} & \multirow{4}{*}{\textbf{0.242}} \\
         & (2) & 1.782 & & & & & \\
         & (3) & 1.732 & & & & & \\
         & (4) & \textbf{2.085} & & & & & \\
        \bottomrule
    \end{tabular}
\end{table*}

\begin{table*}[htbp]
    \centering
    \caption{Evaluation metrics for CEs --- Simulation Case~2 ({mean over 20 base instances}, $S=10$ CEs each). {Bold indicates the top-3 values per metric (all tied values included): highest for \textit{Val}, TIR, \textit{AV/D}, and \textit{AV/P}; lowest for \textit{Dissim} and \textit{Plaus}. See Table~\ref{tab:ta2n1} for abbreviations.}}
    \label{tab:ta2n2}
    \begin{tabular}{llrrrrrr}
        \toprule
        \textbf{Method} & & \textit{Val} & \textit{Dissim} & \textit{Plaus} & TIR & \textit{AV/D} & \textit{AV/P} \\
        \midrule
        M1 (LGB)   & &  4.027 & 11.203 & $1.48\times10^{3}$ &  0.710 & 0.359 & 0.003 \\
        M1 (RF)    & &  4.003 &  8.871 & 676.528 &  0.655 & 0.451 & 0.006 \\
        M1 (MLP)   & &  4.629 &  3.015 &  8.331 & \textbf{1.000} & \textbf{1.535} & \textbf{0.556} \\
        M1 (LR)    & &  2.394 &  3.000 &  8.001 &  0.550 & 0.798 & 0.299 \\
        M2 (Stack) & &  3.455 &  7.601 &  96.218 & 0.410 & 0.455 & 0.036 \\
        \multirow{2}{*}{M3 ($m$=2)} & (1) & \textbf{4.946} & \multirow{2}{*}{\textbf{2.796}} & \multirow{2}{*}{\textbf{6.767}} & \multirow{2}{*}{\textbf{0.880}} & \multirow{2}{*}{\textbf{1.747}} & \multirow{2}{*}{\textbf{0.722}} \\
         & (2) & \textbf{4.826} & & & & & \\
        \multirow{3}{*}{M3 ($m$=3)} & (1) & \textbf{4.705} & \multirow{3}{*}{\textbf{2.781}} & \multirow{3}{*}{\textbf{6.702}} & \multirow{3}{*}{\textbf{0.960}} & \multirow{3}{*}{\textbf{1.591}} & \multirow{3}{*}{\textbf{0.660}} \\
         & (2) & 4.590 & & & & & \\
         & (3) & 3.979 & & & & & \\
        \multirow{4}{*}{M3 ($m$=4)} & (1) & 3.858 & \multirow{4}{*}{\textbf{2.754}} & \multirow{4}{*}{\textbf{7.106}} & \multirow{4}{*}{0.792} & \multirow{4}{*}{1.228} & \multirow{4}{*}{0.476} \\
         & (2) & 3.866 & & & & & \\
         & (3) & 3.460 & & & & & \\
         & (4) & 2.348 & & & & & \\
        \bottomrule
    \end{tabular}
\end{table*}

The parameters were set as $C = 3$ and $\lambda = 2$ (the latter for Methods~1 and~2 only; Method~3 does not use $\lambda$). {The value $C = 3$ was chosen based on the standard deviation of the feature distribution to allow meaningful CEs without excessively large changes. The value $\lambda = 2$ was set to balance validity and proximity for Methods~1 and~2; Table~\ref{tab:sensitivity_lam} reports TIR for $\lambda \in \{1, 2, 5\}$ with $C = 3$ fixed, confirming that TIR is broadly stable across $\lambda$ values for both methods.} For the calculation of \textit{Val}, since $y_t = \infty$, this study simply used the predicted values of $y$, where higher values indicated better evaluations. Moreover, as \textit{Val} and \textit{Dissim} or \textit{Plaus} tended to increase together, the ratios of their average values, \textit{Ave Val}/\textit{Ave Dissim} and \textit{Ave Val}/\textit{Ave Plaus}, were calculated to compare their balance. For Method 3, the average of multiple \textit{Ave Val} values was used as \textit{Ave Val}. The top three lowest values for \textit{Dissim} and \textit{Plaus}, and the top three highest values for \textit{Val}, \textit{TIR}, \textit{Ave Val}/\textit{Ave Dissim} and \textit{Ave Val}/\textit{Ave Plaus}, are highlighted in bold.

\begin{figure}[htbp]
    \centering
    \includegraphics[width=\textwidth]{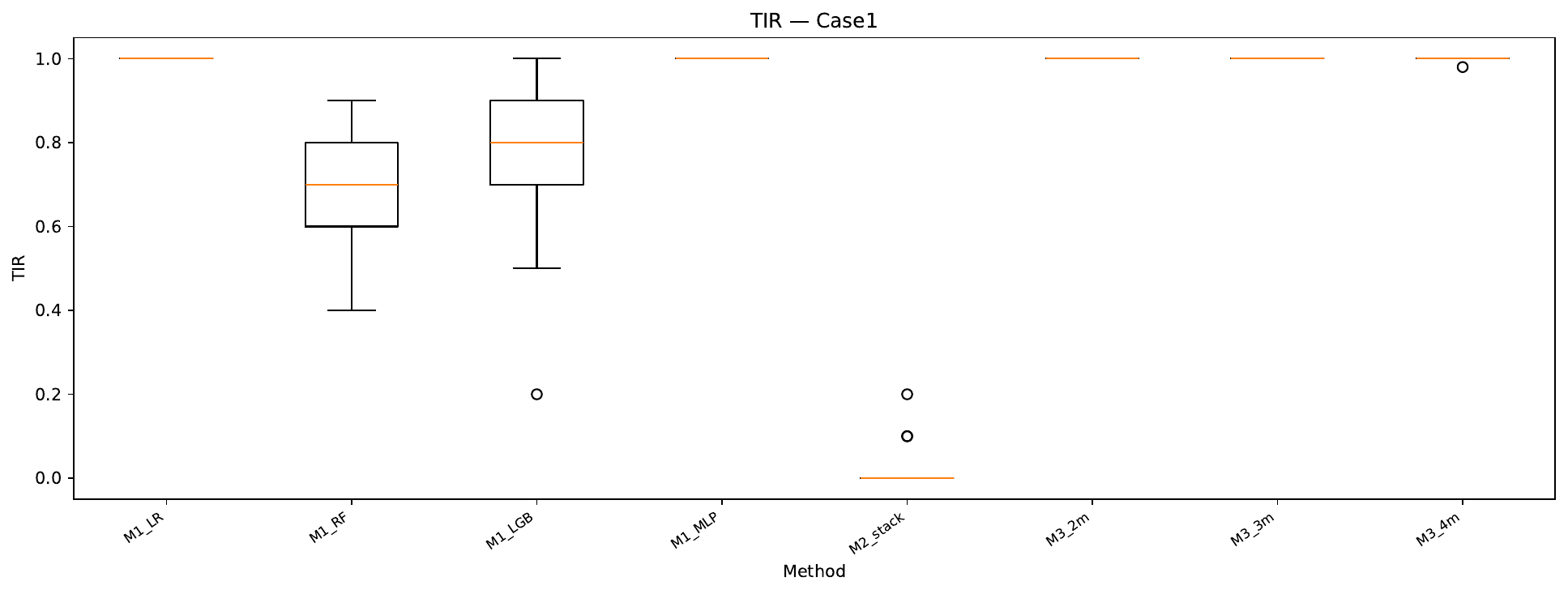}\\[4pt]
    \includegraphics[width=\textwidth]{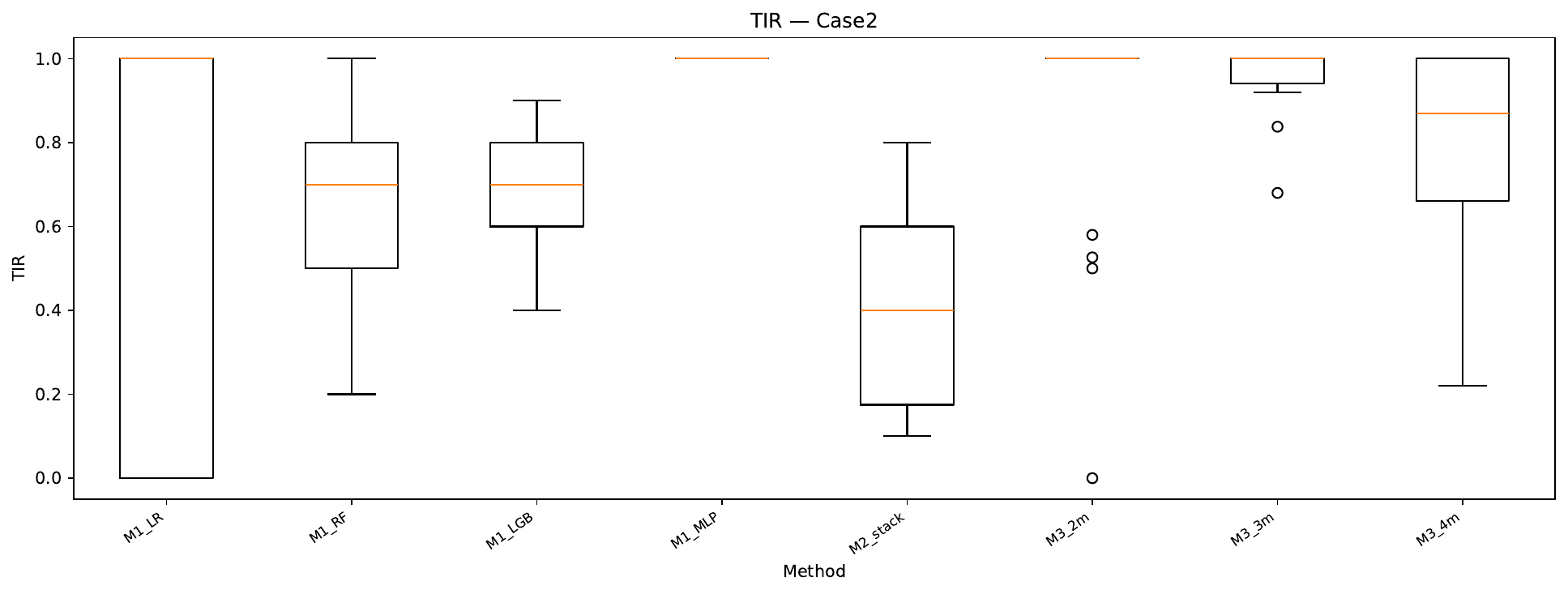}
    \caption{{Boxplots of TIR across 20 base instances for Simulation Case~1 (top) and Case~2 (bottom).}}
    \label{fig:sim_boxplot_tir}
\end{figure}

\begin{figure}[htbp]
    \centering
    \includegraphics[width=\textwidth]{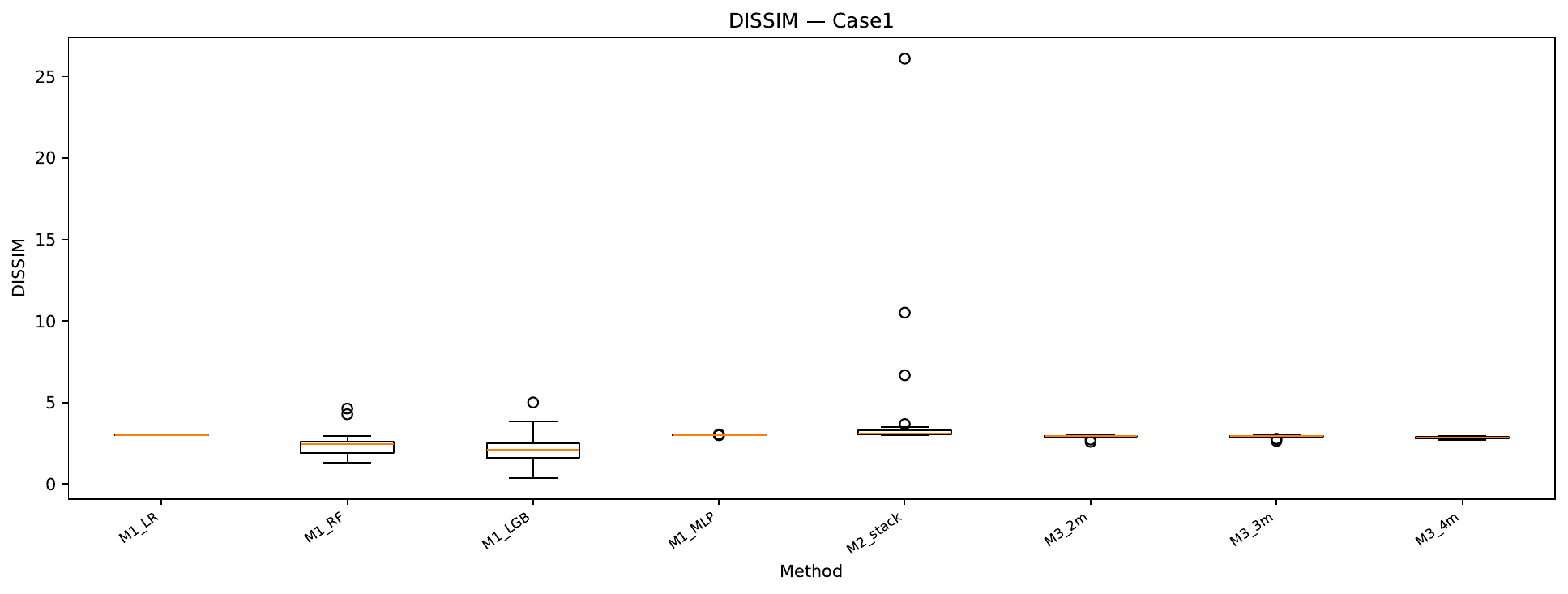}\\[4pt]
    \includegraphics[width=\textwidth]{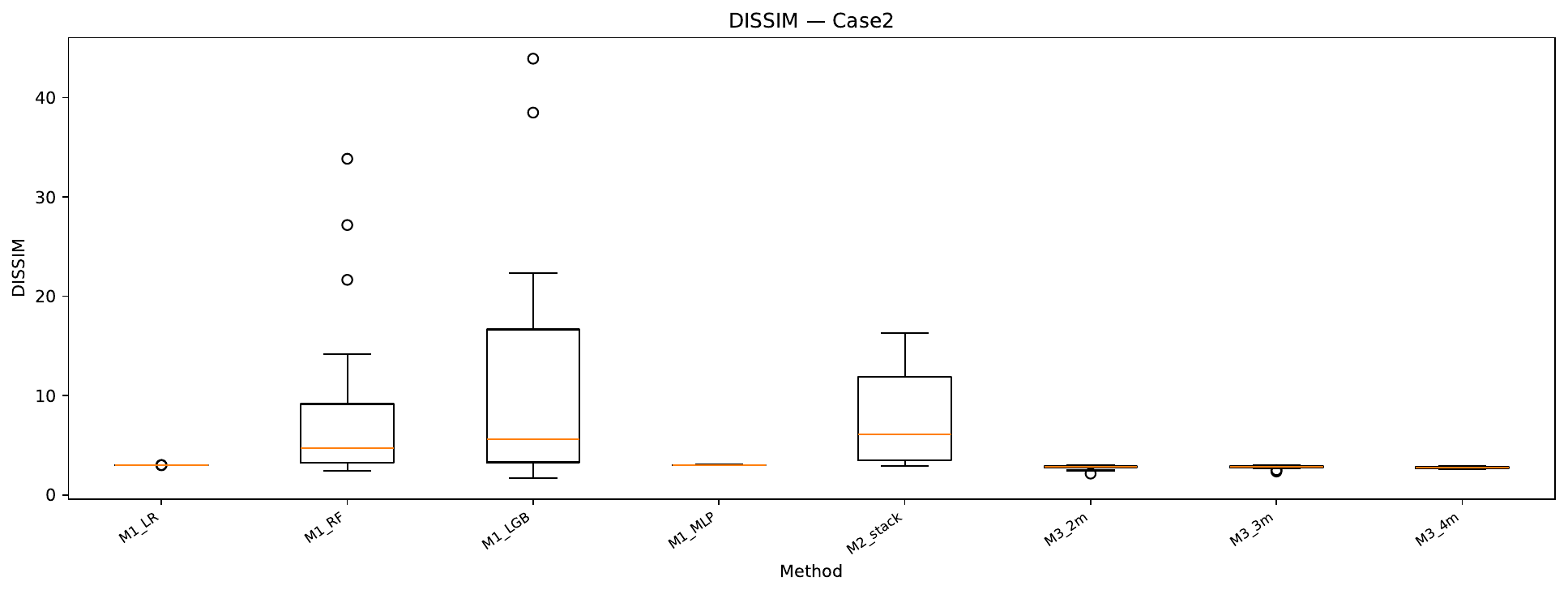}
    \caption{{Boxplots of Dissim across 20 base instances for Simulation Case~1 (top) and Case~2 (bottom).}}
    \label{fig:sim_boxplot_dissim}
\end{figure}

{In Case 1 (Table~\ref{tab:ta2n1}), which represents the full model-multiplicity setting, Method~3 achieved TIR~=~1.000 for $m \in \{2, 3\}$ and TIR~=~0.999 for $m=4$, with compact Dissim (M3\_3m: 2.90; M3\_4m: 2.84) and Plaus (M3\_3m: 7.80; M3\_4m: 7.37). M1~(LGB) achieved the smallest Dissim (2.193) but at the cost of lower TIR (0.765); M1~(RF) achieved the smallest Plaus (6.647) at the cost of lower TIR (0.705). Notably, Method~2 (stacking) achieved TIR~=~0.020---essentially zero---indicating that the stacking ensemble fails to generalise to the true G function even under full model multiplicity. Method~3 achieved the best TIR while maintaining the smallest Dissim and Plaus among all methods with TIR~=~1.000. Figure~\ref{fig:sim_boxplot_tir} confirms that M3\_2m and M3\_3m achieve TIR~=~1.000 with zero variance across all 20 instances, and M3\_4m achieves TIR~=~0.999 with near-zero variance, whereas M2\_stack degenerates to near-zero TIR in every case. Figure~\ref{fig:sim_boxplot_dissim} shows that M2\_stack exhibits large Dissim variance, while M3 variants maintain compact, low-variance distributions.}

{In Case 2 (Table~\ref{tab:ta2n2}), which represents the valid model-multiplicity setting for nonlinear models, Method~3 again achieved the smallest \textit{Dissim} (2.75--2.80) and \textit{Plaus} (6.70--7.11) across all methods. Method~1~(MLP) achieved the highest TIR~=~1.000, reflecting that the Case~2 true function is well captured by MLP when optimised alone. Method~3 achieved TIR in the range 0.792--0.960, providing solid robustness while simultaneously satisfying the three nonlinear models selected by the top-$m$ criterion (LGB, RF, MLP). Method~2 (stacking) showed the lowest TIR~=~0.410, confirming that a stacking ensemble that includes the weaker LR model fails to generalise under the Case~2 true function. The contrast highlights that the proposed method's explicit multi-objective structure, combined with MSE-based model selection, is more effective than stacking under model multiplicity. Figure~\ref{fig:sim_boxplot_tir} confirms that M3 variants maintain consistently high TIR with moderate variance, while M2\_stack shows low median TIR and high instance-to-instance variance. Figure~\ref{fig:sim_boxplot_dissim} reveals that M1\_LGB and M1\_RF produce highly variable and large Dissim values in Case~2 (medians $\approx5.6$ and $\approx4.7$, with individual cases reaching above 40) due to the instability of gradient-based optimisation for tree-based models, whereas M3 variants remain compact across all instances (medians $\approx2.8$).}

\begin{table}[htbp]
    \centering
    \caption{Sensitivity of TIR to the regularisation weight $\lambda$ for Methods~1 and~2 ($C = 3$ fixed, averaged over 20 base instances independently sampled from those in Tables~\ref{tab:ta2n1}--\ref{tab:ta2n2}). TIR is stable across $\lambda \in \{1, 2, 5\}$ for smooth-function models (LR, MLP) and near-stable for tree-based models (RF, LGB).}
    \label{tab:sensitivity_lam}
    \begin{tabular}{llccc|ccc}
        \toprule
        & & \multicolumn{3}{c|}{\textbf{Case~1 TIR}} & \multicolumn{3}{c}{\textbf{Case~2 TIR}} \\
        \textbf{Method} & & $\lambda=1$ & $\lambda=2$ & $\lambda=5$ & $\lambda=1$ & $\lambda=2$ & $\lambda=5$ \\
        \midrule
        M1 (LR)    & & 1.000 & 1.000 & 1.000 & 0.450 & 0.450 & 0.450 \\
        M1 (RF)    & & 0.790 & 0.740 & 0.755 & 0.675 & 0.620 & 0.610 \\
        M1 (LGB)   & & 0.800 & 0.715 & 0.770 & 0.710 & 0.670 & 0.715 \\
        M1 (MLP)   & & 1.000 & 1.000 & 1.000 & 1.000 & 0.995 & 0.995 \\
        M2 (Stack) & & 0.045 & 0.040 & 0.030 & 0.365 & 0.365 & 0.410 \\
        \bottomrule
    \end{tabular}
\end{table}

Additionally, we examined the impact of $C$ on TIR. {Table~\ref{tab:sensitivity_c} reports TIR for Method~3 (3-model) across $C \in \{1, 3, 5\}$ for both simulation cases. Since Method~3 does not incorporate $\lambda$ in its objective, TIR is invariant to $\lambda$ by construction (all entries within each $C$ row are identical across $\lambda$ columns). In Case~1, TIR was 0.912 at $C=1$ and reached 1.000 at $C=3$ and $C=5$, suggesting that a sufficient proximity bound is needed to achieve full robustness. In Case~2, TIR increased monotonically with $C$ (0.646, 0.800, 0.820), confirming that $C$ is the key parameter governing robustness.}

\begin{table}[htbp]
    \centering
    \caption{Sensitivity of TIR to the proximity bound $C$ (Method~3, 3-model, averaged over 20 base instances). Method~3 does not use $\lambda$; $C$ is the primary problem-setting parameter governing TIR. Sensitivity to algorithmic parameters (pop, gen) is examined in Table~\ref{tab:sensitivity_nsga2}.}
    \label{tab:sensitivity_c}
    \begin{tabular}{ccc}
        \toprule
        $C$ & \textbf{Case~1 TIR} & \textbf{Case~2 TIR} \\
        \midrule
        1 & 0.912 & 0.646 \\
        3 & 1.000 & 0.800 \\
        5 & 1.000 & 0.820 \\
        \bottomrule
    \end{tabular}
\end{table}

{Table~\ref{tab:sensitivity_nsga2} reports the sensitivity of TIR and \textit{Dissim} to NSGA-II hyperparameters---population size (\textit{pop}) and number of generations (\textit{gen})---for Method~3 ($m$=3), averaged over the same 20 base instances as the main experiment. In Case~1, TIR~=~1.000 is achieved for all settings except the minimal combination (pop=20, gen=20; TIR~=~0.989), confirming that results are broadly robust to hyperparameter choice. In Case~2, TIR ranges from 0.962 to 0.997 across all settings; the larger standard deviations reflect instance-level variability driven by a small number of structurally difficult base instances rather than a systematic trend with pop or gen. Increasing \textit{pop} and \textit{gen} does not consistently improve TIR, but does increase the mean number of Pareto-optimal solutions (from $\approx7$ at pop=20/gen=20 to $\approx70$ at pop=100/gen=50 in Case~1), offering the user more diverse CE options at higher computational cost. The paper's default setting (pop=50, gen=30, marked with $\dagger$) lies in the stable region for both cases.}

\begin{table}[htbp]
    \centering
    \caption{{Sensitivity of TIR and \textit{Dissim} to NSGA-II hyperparameters for Method~3 ($m$=3, $C=3$, averaged over 20 base instances). $\dagger$~marks the default setting for Experiment~1 (simulation; pop=50, gen=30); Experiment~3 (real data) uses gen=20. Case~1 TIR is 1.000 for all settings except pop=20/gen=20. Case~2 values are reported as mean~$\pm$~SD to reflect instance-level variability. \textit{Dissim} is stable across all settings ($\approx$2.75--2.95). Sensitivity analysis for real data remains future work (Section~\ref{sec5}).}}
    \label{tab:sensitivity_nsga2}
    \begin{tabular}{cccccc}
        \toprule
        & & \multicolumn{2}{c}{\textbf{Case~1}} & \multicolumn{2}{c}{\textbf{Case~2}} \\
        \cmidrule(lr){3-4}\cmidrule(lr){5-6}
        \textbf{pop} & \textbf{gen} & \textbf{TIR} & \textbf{Dissim} & \textbf{TIR (mean$\pm$SD)} & \textbf{Dissim} \\
        \midrule
         20 &  20 & 0.989 & 2.750 & $0.997 \pm 0.011$ & 2.760 \\
         20 &  30 & 1.000 & 2.861 & $0.990 \pm 0.044$ & 2.835 \\
         20 &  50 & 1.000 & 2.921 & $0.980 \pm 0.062$ & 2.895 \\
         50 &  20 & 1.000 & 2.867 & $0.963 \pm 0.099$ & 2.799 \\
         $50^{\dagger}$ & $30^{\dagger}$ & 1.000 & 2.922 & $0.975 \pm 0.092$ & 2.869 \\
         50 &  50 & 1.000 & 2.950 & $0.978 \pm 0.080$ & 2.918 \\
        100 &  20 & 1.000 & 2.909 & $0.966 \pm 0.101$ & 2.840 \\
        100 &  30 & 1.000 & 2.934 & $0.964 \pm 0.105$ & 2.893 \\
        100 &  50 & 1.000 & 2.954 & $0.962 \pm 0.099$ & 2.931 \\
        \bottomrule
    \end{tabular}
\end{table}

In summary, in Case~2, Method~3 achieved the smallest \textit{Dissim} and \textit{Plaus} among all methods. In Case~1, M1~(LGB) achieved the smallest Dissim (2.193) and M1~(RF) achieved the smallest Plaus (6.647), but both at the cost of lower TIR ($<$1.000), whereas Method~3 maintained TIR~$\geq$~0.999 (1.000 for $m \in \{2,3\}$, 0.999 for $m=4$). In Case~1 (full model-multiplicity), Method~3 achieved TIR~$\geq$~0.999 with competitive proximity, while Method~2 (stacking) collapsed to TIR~=~0.020, demonstrating that multi-objective optimisation substantially outperforms stacking under full model multiplicity. In Case~2 (nonlinear model-multiplicity), Method~3 achieved TIR in the range 0.792--0.960, outperforming Method~2 (TIR~=~0.410). Across both cases, Method~3 consistently yields the most reliable and proximity-efficient CEs. {The lower average \textit{Val} of Method~3 ($\approx$1.752 for $m$=3) relative to the best single-model approach, M1~(MLP) (1.997), in Case~1 is an expected consequence of the multi-objective formulation: optimizing simultaneously across multiple models prevents any single model's prediction from being maximized at the expense of the others. This Val--robustness trade-off is the design intent of the proposed method, and is outweighed by the substantial gains in proximity and plausibility.}

In Case~1 (Figure~\ref{fig:sim_boxplot_tir}), M3\_2m and M3\_3m achieve TIR~=~1.000 with zero variance; M3\_4m achieves TIR~=~0.999 with near-zero variance. M2\_stack collapses to near-zero TIR across all instances. In Case~2, M1\_MLP achieves the highest TIR (1.000) with zero variance; Method~3 variants maintain high TIR (0.792--0.960) with moderate variance, while M2\_stack shows the lowest median TIR (0.410) and high variance. Regarding Dissim (Figure~\ref{fig:sim_boxplot_dissim}), in Case~1, M3 variants achieve compact distributions; M2\_stack exhibits high variance reflecting stacking generalisation failure. In Case~2, M3 variants achieve the smallest Dissim (medians $\approx2.8$); M1\_LGB and M1\_RF exhibit highly variable and large Dissim values (medians $\approx5.6$ and $\approx4.7$, with individual cases exceeding 40) due to the instability of gradient-based optimisation for tree-based models.

{
\subsection{Experiment 2: Benchmark Comparison}\label{subsec7}

{We compare the proposed method against two established single-model CE baselines. \textbf{Wachter-style optimization} \cite{wachter2017} generates a CE by minimising a proximity-penalised validity loss for a single surrogate model. \textbf{DiCE} \cite{mothilal2020} generates a diverse set of CEs for a single surrogate model using a genetic algorithm; CEs are post-processed to satisfy proximity constraints. Both baselines optimise for a single model only, in contrast to the proposed method which simultaneously optimises across all selected models.}

The three methods compared are: (1)~\textbf{Wachter}: proximity-constrained optimization via SLSQP targeting the single best-MSE model ($S=10$ random restarts, $\lambda=2$, $C=3$), with post-processing to clip CEs exceeding $C=3$; \textit{Val} is computed using that model's prediction; (2)~\textbf{DiCE}: genetic algorithm backend with the same best-MSE model (20 CFs, proximity weight 0.5, diversity weight 1.0), with post-processing to clip CEs exceeding $C=3$; \textit{Val} is computed similarly; (3)~\textbf{Proposed (Method~3)}: the MOO-based method using the top three models by MSE (NSGA-II, $\text{pop}=50$, $\text{gen}=30$, SBX prob$=0.9$, $\eta=20$, PM $\eta=20$, $C=3$); individual model \textit{Val} values are reported per row, and \textit{AV/D} and \textit{AV/P} use the average \textit{Val} across the three selected models (consistent with Experiment~1).
The same twenty base instances as in Experiment~1 (Section~\ref{subsec5}) were used for each simulation case, ensuring direct comparability of results. {The same seeds as Experiment~1 were used: data generation and model training used seed~42; base-case sampling used seed~0 (Case~1) and seed~99 (Case~2); NSGA-II used seed~1.} All other hyperparameters follow those described in Section~\ref{subsec3}.

Table~\ref{tab:benchmark_sim} reports the mean values of \textit{Val}, \textit{Dissim}, \textit{Plaus}, and TIR for each simulation case.

\begin{table*}[htbp]
\caption{{Benchmark comparison on simulation data (mean over 20 base instances). Bold indicates the best value per metric per case (ties included); since only three methods are compared, the single best value is highlighted rather than the top-3 used in Tables~\ref{tab:ta2n1}--\ref{tab:ta2n2}: highest for \textit{Val}, TIR, \textit{AV/D}, and \textit{AV/P}; lowest for \textit{Dissim} and \textit{Plaus}. See Table~\ref{tab:ta2n1} for abbreviations.}}
\label{tab:benchmark_sim}
\centering
\begin{tabular}{lllrrrrrrr}
\hline
Case & Method & & Val & Dissim & Plaus & TIR & \textit{AV/D} & \textit{AV/P} \\
\hline
\multirow{5}{*}{Case 1}
 & Wachter         & & 1.780 & 3.000 & 8.434 & \textbf{1.000} & 0.593 & 0.211 \\
 & DiCE            & & 1.363 & 2.998 & 8.134 & 0.962 & 0.455 & 0.168 \\
 & \multirow{3}{*}{Proposed (M3)} & (1) & 1.602 & \multirow{3}{*}{\textbf{2.904}} & \multirow{3}{*}{\textbf{7.799}} & \multirow{3}{*}{\textbf{1.000}} & \multirow{3}{*}{\textbf{0.603}} & \multirow{3}{*}{\textbf{0.225}} \\
 & & (2) & \textbf{1.836} & & & & & \\
 & & (3) & 1.818 & & & & & \\
\hline
\multirow{5}{*}{Case 2}
 & Wachter         & & 3.266 & \textbf{2.609} & \textbf{6.623} & 0.635 & 1.252 & 0.493 \\
 & DiCE            & & 3.166 & 2.974 & 7.565 & 0.632 & 1.065 & 0.419 \\
 & \multirow{3}{*}{Proposed (M3)} & (1) & \textbf{4.705} & \multirow{3}{*}{2.781} & \multirow{3}{*}{6.702} & \multirow{3}{*}{\textbf{0.960}} & \multirow{3}{*}{\textbf{1.591}} & \multirow{3}{*}{\textbf{0.660}} \\
 & & (2) & 4.590 & & & & & \\
 & & (3) & 3.979 & & & & & \\
\hline
\end{tabular}
\end{table*}

In Case~1, both the proposed method and Wachter achieved the highest TIR ($1.000$), while DiCE attained $0.962$. The proposed method achieved the best proximity: \textit{Dissim}~$=2.904$ and \textit{Plaus}~$=7.799$, compared to Wachter ($3.000$, $8.434$) and DiCE ($2.998$, $8.134$). Among individual model Val values for the proposed method, model~2 attained the highest Val overall ($1.836$), exceeding Wachter ($1.780$) and DiCE ($1.363$). In Case~2, the proposed method's model~1 achieved the highest Val ($4.705$), substantially outperforming Wachter ($3.266$) and DiCE ($3.166$), and the highest TIR ($0.960$) compared to Wachter ($0.635$) and DiCE ($0.632$). Wachter achieved the smallest \textit{Dissim} ($2.609$) and \textit{Plaus} ($6.623$), with the proposed method competitive ($2.781$, $6.702$). The proposed method achieved the highest \textit{AV/P} and \textit{AV/D} in both cases, demonstrating that multi-objective optimization across models produces CEs that more reliably improve the true outcome while maintaining competitive proximity.
}

\subsection{Experiment 3: Real Data}\label{subsec6}

Two real-world datasets evaluate the method's applicability across different feature types. Experiment~3.1 uses a continuous-feature dataset based on web search trends to demonstrate that the framework handles continuous variables with explicit bounds. Experiment~3.2 applies the method to an educational intervention dataset with predominantly binary features and equality constraints, demonstrating the method's constraint-handling flexibility.

\subsubsection{Experiment 3.1: Google Trends Data}\label{subsec6a}

{This experiment uses weekly Google Trends search interest data (2019-12-29 to 2025-02-23; $n = 270$ weekly observations) to demonstrate that the proposed method operates on continuous, bounded features. The target variable is the search interest for ``sake'' ($y \in [0, 100]$), and the five features are ``washoku'' (Japanese cuisine), ``nabe'' (Japanese hot pot), ``sushi'', ``ramen'', and ``izakaya'' (all $\in [0, 100]$), each independently normalized to retain its own 0--100 scale. The CE task is to identify what changes to complementary food-category search trends would increase predicted sake interest---analogous to a marketing intervention where a brand raises product awareness by influencing related category trends.}

{The relationship between these food-category search trends and sake interest is likely to involve nonlinear seasonal patterns and feature interactions (e.g., nabe searches peak sharply in winter, a season strongly associated with sake consumption, while the combination of izakaya and sushi trends may jointly predict sake interest more strongly than either alone). Tree-based models (RF, LGB) are therefore included to capture such nonlinear dependencies; since these models are opaque, the CE framework provides interpretable recommendations for what changes to search trends would increase sake interest.}

{The four models from Section~\ref{subsec5} were trained on a 70/30 split ($n_{\text{train}} = 189$, $n_{\text{test}} = 81$) repeated 20 times. Table~\ref{tab:gt_accuracy} reports the MSE mean $\pm$ SD values. Model~4 (MLP, MSE$\,{=}\,143.166 \pm 13.801$) performed substantially worse than the other three, likely due to the limited training size relative to model complexity; accordingly, the top-3 models (LR, RF, LGB) were selected for Method~3. Their mean~$\pm$~SD ranges overlapped (max lower bound~$= 11.398$, min upper bound~$= 19.987$), justifying the model-multiplicity setting with $m = 3$.}

\begin{table}[htbp]
    \centering
    \caption{{Model accuracy for Google Trends data (MSE mean $\pm$ SD over 20 splits). Top-3 base models (LR, RF, LGB) have overlapping mean $\pm$ SD ranges and are selected for Method~3; MLP is excluded. Stacking is reserved for Method~2.}}
    \begin{tabular}{lc}
        \hline
        \textbf{Model} & \textbf{MSE (mean $\pm$ SD)} \\
        \hline
        Model 1 (LR)   & $14.175 \pm 5.812$ \\
        Model 2 (RF)   & $15.958 \pm 4.560$ \\
        Model 3 (LGB)  & $15.727 \pm 5.129$ \\
        Model 4 (MLP)  & $143.166 \pm 13.801$ \\
        Stacking Model & $19.240 \pm 3.962$ \\
        \hline
    \end{tabular}
    \label{tab:gt_accuracy}
\end{table}

{Forty base cases were randomly selected from the test set {(base-case selection: seed~0; SLSQP initial values for Methods~1 and~2: seed~42; NSGA-II: seed~1)}. The proximity bound was set to $C = 5.0$ (permitting shifts of up to 5 points on the 0--100 scale); for Methods~1 and~2, the regularisation weight was set to $\lambda = 5.0$ (Method~3 does not use $\lambda$); and all five features were constrained to $[0, 100]$ via upper- and lower-bound inequality constraints. Table~\ref{tab:gt_ce} reports the CE evaluation metrics averaged over 40 base cases ($S = 20$ CEs each). TIR is not computable for real data.}

\begin{table}[htbp]
    \centering
    \caption{{CE evaluation metrics for Google Trends data (mean over 40 base cases, $S = 20$ CEs each). Bold indicates the top-3 values per metric (all tied values included): highest \textit{Val}, $\Delta$\textit{Val}, \textit{AV/D}, and \textit{AV/P}; lowest \textit{Dissim} and \textit{Plaus}. See Table~\ref{tab:ta2n1} for abbreviations.}}
    \label{tab:gt_ce}
    \begin{tabular}{llrrrrrr}
        \toprule
        \textbf{Method} & & \textit{Val} & $\Delta$\textit{Val} & \textit{Dissim} & \textit{Plaus} & \textit{AV/D} & \textit{AV/P} \\
        \midrule
        M1 (LR)    & & 69.426 & $+2.301$ & 5.000 & 22.749 & 13.885 & 3.052 \\
        M1 (RF)    & & 69.146 & $+2.021$ & \textbf{2.823} & \textbf{10.475} & \textbf{24.494} & \textbf{6.601} \\
        M1 (LGB)   & & 69.372 & $+2.247$ & \textbf{2.723} & \textbf{10.253} & \textbf{25.476} & \textbf{6.766} \\
        M1 (MLP)   & & \textbf{74.054} & $\mathbf{+6.929}$ & 4.612 & 19.818 & 16.057 & 3.737 \\
        M2 (Stack) & & 68.977 & $+1.852$ & \textbf{4.220} & \textbf{17.411} & \textbf{16.345} & \textbf{3.962} \\
        \multirow{3}{*}{M3 ($m$=3)} & (1) & 68.904 & $+1.779$ & \multirow{3}{*}{4.822} & \multirow{3}{*}{21.188} & \multirow{3}{*}{14.458} & \multirow{3}{*}{3.290} \\
         & (2) & \textbf{70.140} & $\mathbf{+3.015}$ & & & & \\
         & (3) & \textbf{70.108} & $\mathbf{+2.983}$ & & & & \\
        \bottomrule
    \end{tabular}
\end{table}

{All methods achieved positive $\Delta\textit{Val}$, confirming that the CE framework successfully identifies search-trend changes that increase predicted sake interest. Method~3 achieved positive $\Delta\textit{Val}$ across all three selected models ($+1.779$, $+3.015$, $+2.983$ for Models~1--3, respectively), outperforming stacking ($+1.852$) on average. M1~(MLP) achieved the highest $\Delta\textit{Val}$ overall ($+6.929$), though it optimizes a single model only. M1~(LGB) achieved the smallest Dissim ($2.723$) and Plaus ($10.253$), but also at the cost of single-model optimization. Crucially, all three per-model $\Delta\textit{Val}$ values are positive ($+1.779$, $+3.015$, $+2.983$), which is direct evidence of robustness under model multiplicity: the generated CEs improve the predicted outcome simultaneously across all selected models. Method~1, by contrast, optimises a single model and provides no such guarantee for the others.}

\begin{figure}[htbp]
    \centering
    \includegraphics[width=0.8\textwidth]{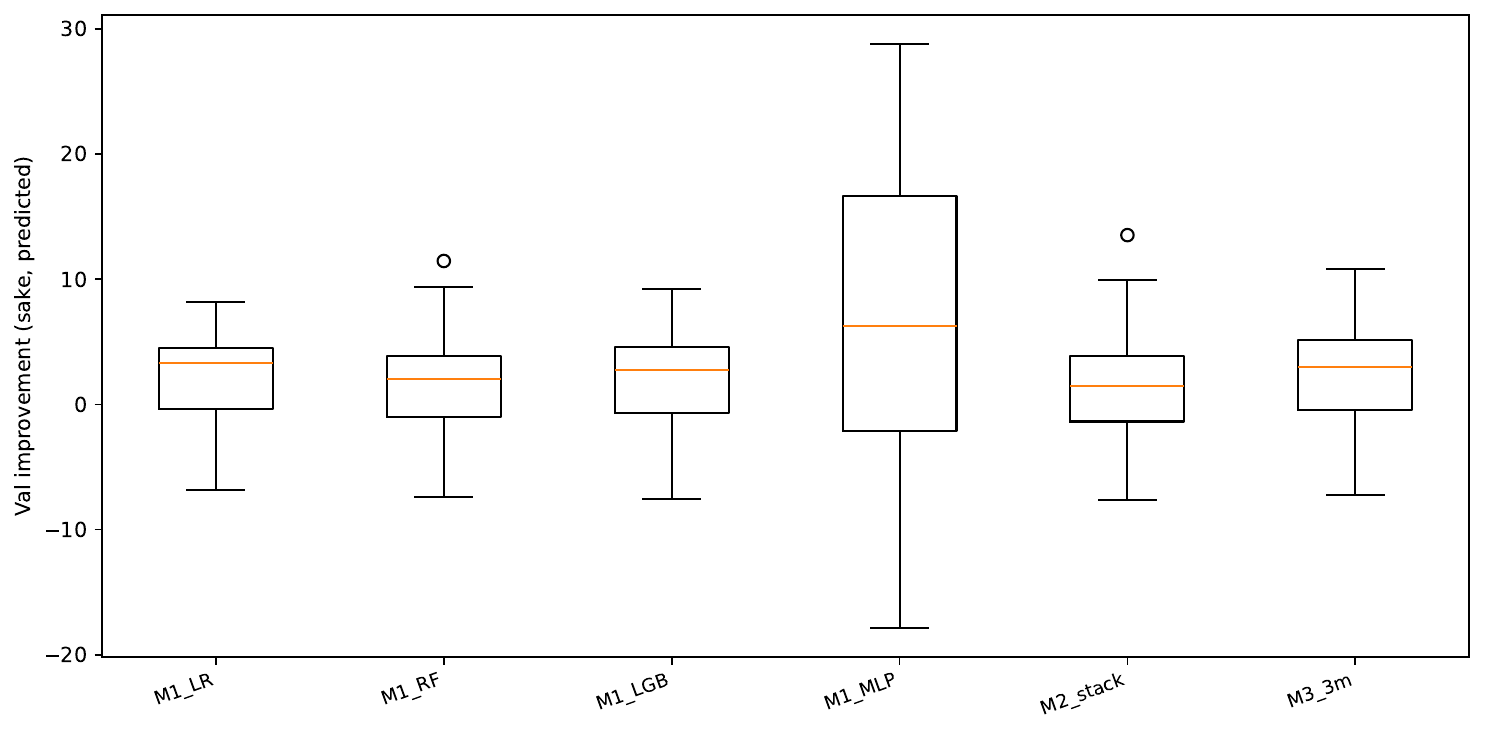}
    \caption{{Boxplot of Val improvement ($\Delta\text{Val}$) for each method on Google Trends data (40 base cases). Method labels correspond to: M1\_LR, M1\_RF, M1\_LGB, M1\_MLP (Method~1 targeting each single model), M2\_stack (Method~2), M3\_3m (Method~3, three-model MOO). Higher values indicate greater improvement over the base instance.}}
    \label{fig:gt_boxplot_val_improv}
\end{figure}

\begin{figure}[htbp]
    \centering
    \includegraphics[width=0.8\textwidth]{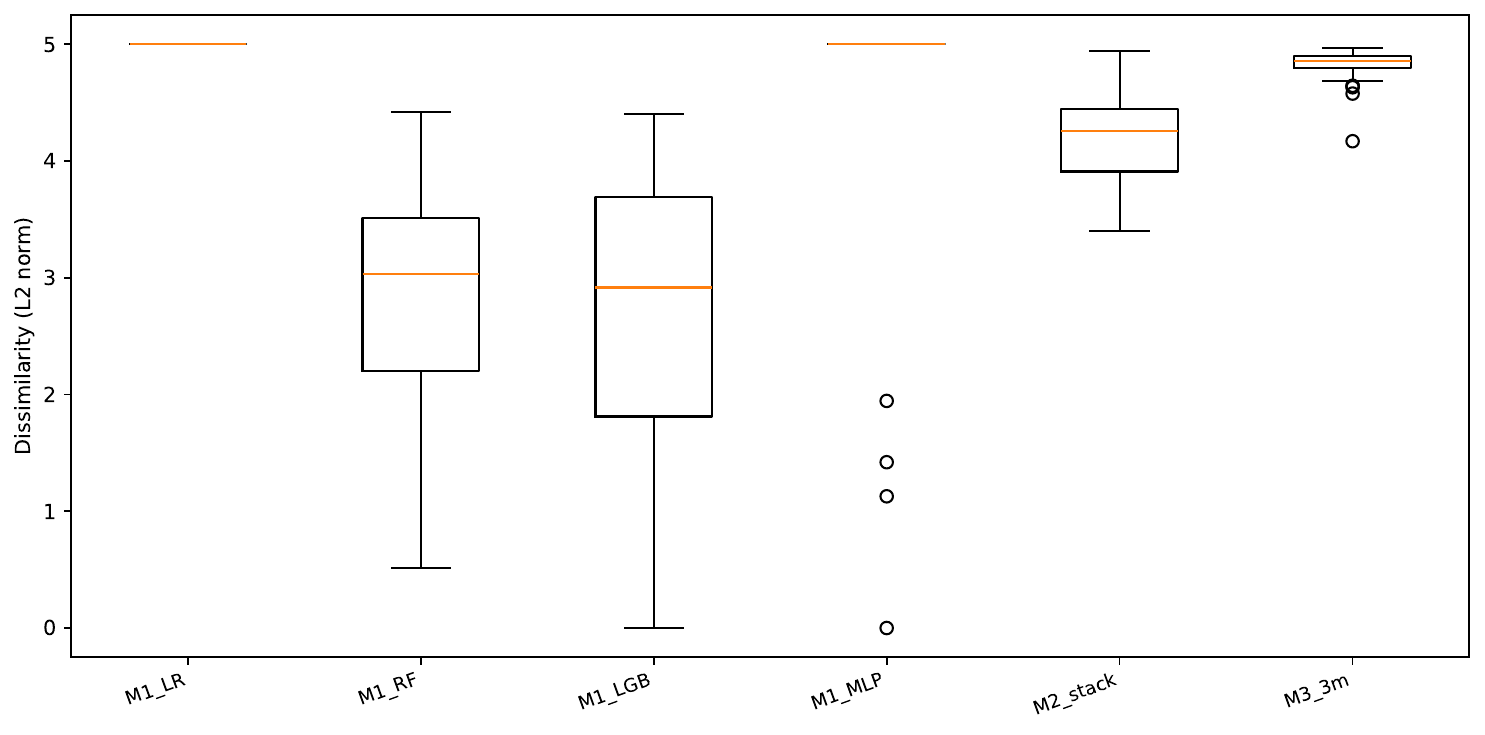}
    \caption{{Boxplot of Dissim for each method on Google Trends data (40 base cases). Lower values indicate that the CE requires smaller changes from the base instance. Methods hitting the proximity ceiling ($C = 5.0$) show a degenerate distribution at 5.0.}}
    \label{fig:gt_boxplot_dissim}
\end{figure}

{Figure~\ref{fig:gt_boxplot_val_improv} and Figure~\ref{fig:gt_boxplot_dissim} show the distribution of Val improvement and Dissim across the 40 base cases. Method~3 achieves mostly positive Val improvement, confirming robust multi-model performance. In terms of Dissim, M1~(LR) and M1~(MLP) both hit the proximity ceiling ($\text{Dissim} = 5.0$) in virtually all cases; M1~(MLP) exhibits several below-ceiling outliers (down to $\approx 0$), indicating a small number of cases in which the base instance was already close to the target. M1~(LGB) and M1~(RF) achieve lower Dissim ($2.72$ and $2.82$, respectively) at the cost of optimizing a single model only. M2~(Stack) shows Dissim values around $\approx 4.2$, while M3 is concentrated near the ceiling ($\approx 4.8$), reflecting the constraint that simultaneously satisfying multiple model objectives tends to use most of the available proximity budget.}

\begin{figure}[htbp]
    \centering
    \includegraphics[width=0.7\textwidth]{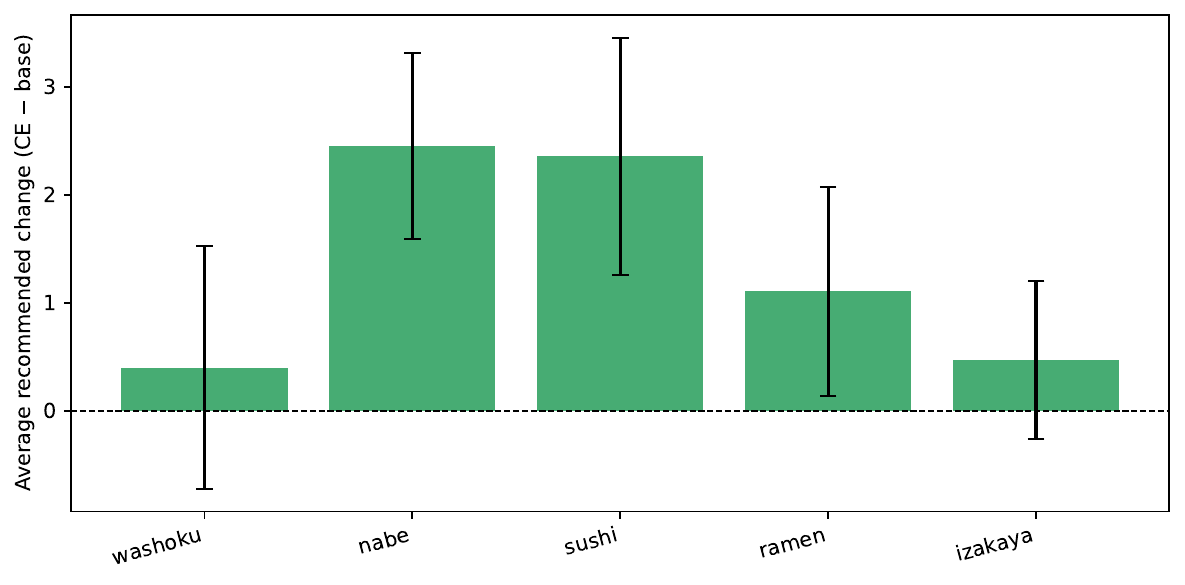}
    \caption{{Average recommended change per feature (CE minus base value), aggregated as the mean of per-case means over 40 base cases. All features receive positive recommendations; error bars indicate standard deviation across the 40 cases.}}
    \label{fig:gt_fig2}
\end{figure}

\begin{figure}[htbp]
    \centering
    \includegraphics[width=0.8\textwidth]{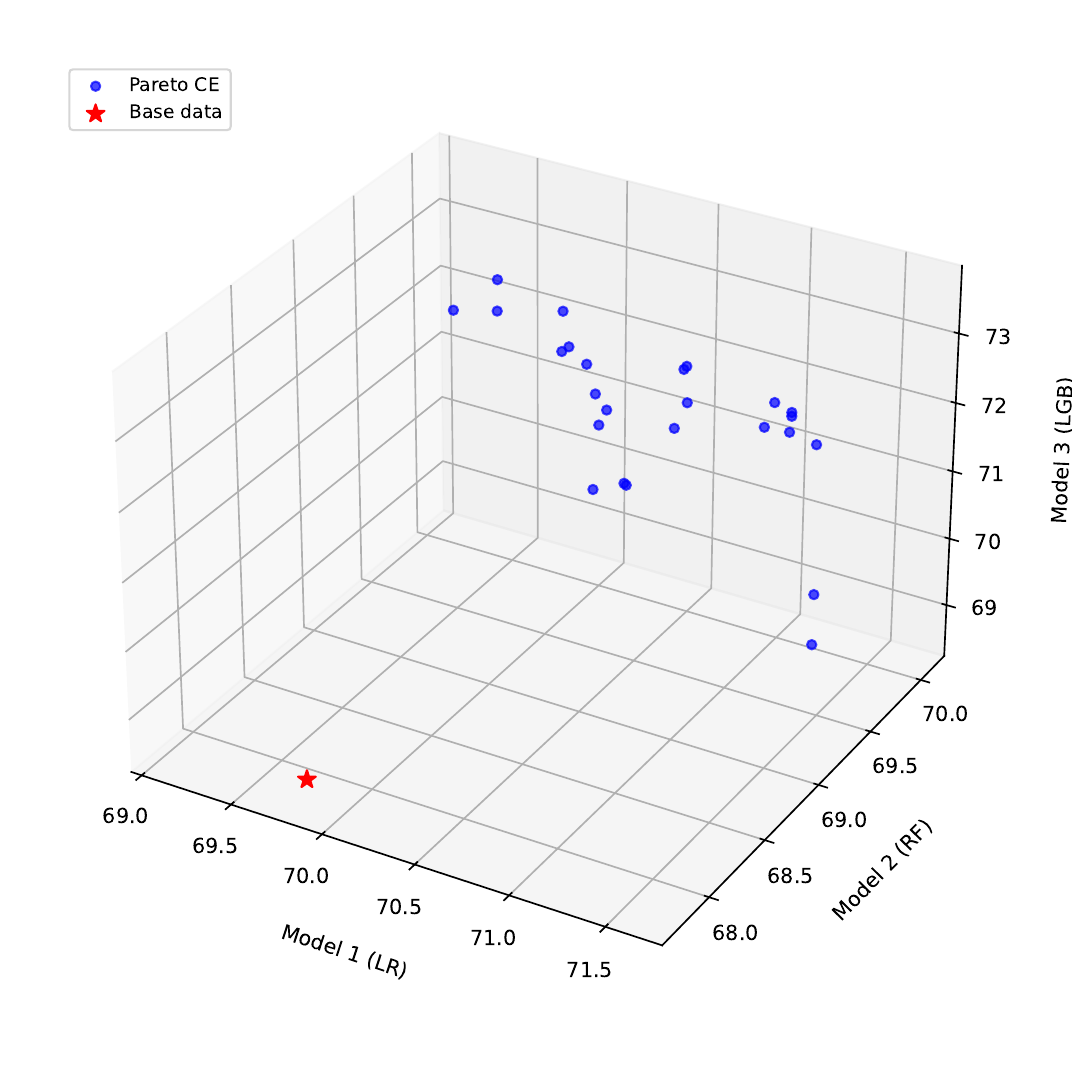}
    \caption{{Pareto front for a representative base case (Google Trends). Each blue point is a CE solution on the Pareto front; the red star is the base data. The three axes correspond to the predicted sake interest under Model~1 (LR), Model~2 (RF), and Model~3 (LGB). The spread of points illustrates the diversity of trade-off solutions.}}
    \label{fig:gt_fig3}
\end{figure}

\begin{figure}[htbp]
    \centering
    \includegraphics[width=0.65\textwidth]{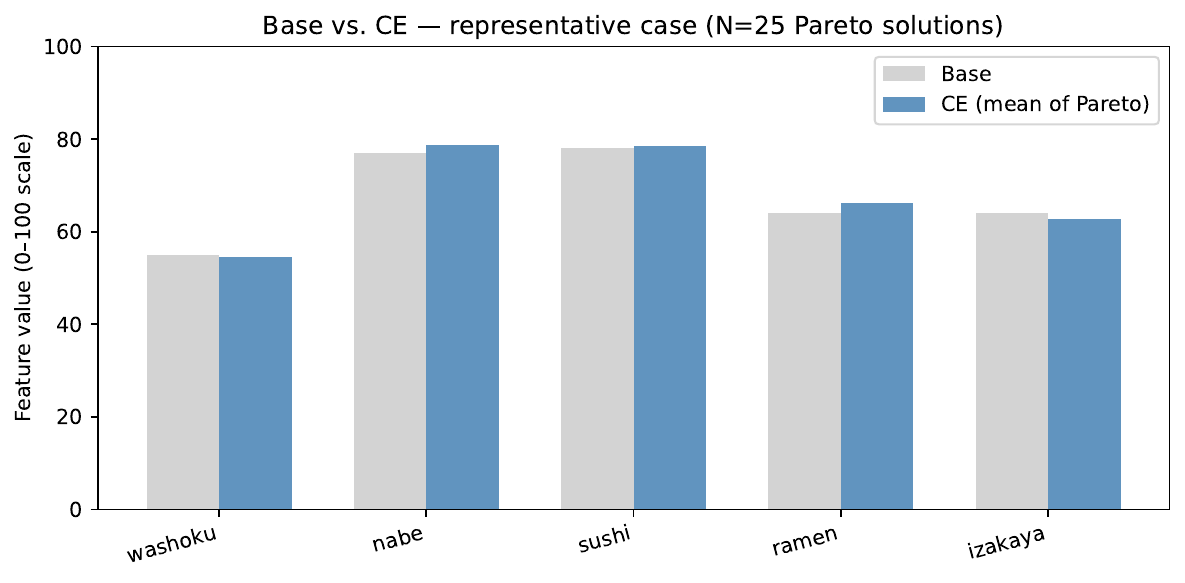}
    \caption{{Base feature values compared with mean CE values for the representative base case (Method~3). Each pair of bars shows the base value (gray) and the mean CE value across all solutions on the Pareto front (blue) for each feature.}}
    \label{fig:gt_fig_base_ce}
\end{figure}

{Figure~\ref{fig:gt_fig2} shows the average recommended change (CE minus base value) per feature across 40 base cases. All five features receive positive recommendations, with \textit{nabe} ($\approx+2.5$) and \textit{sushi} ($\approx+2.4$) receiving the largest increases, followed by \textit{ramen} ($\approx+1.1$), and \textit{washoku} and \textit{izakaya} receiving smaller positive changes ($\approx+0.5$ each). This pattern suggests that increasing search activity across all food-category features---particularly \textit{nabe} and \textit{sushi}, which have strong seasonal association with sake---is predicted to increase sake interest across all three models. Figure~\ref{fig:gt_fig3} shows the Pareto front for a representative base case. The blue points (CE solutions) are distributed above and away from the base data (red star) in the three-dimensional model prediction space, confirming that the proposed method generates CEs that improve predicted sake interest across multiple models. The spread of the Pareto cloud illustrates genuine trade-offs between model predictions: some CEs achieve higher LGB predictions at the cost of slightly lower RF predictions, and vice versa, reflecting the model-multiplicity structure of the problem. Figure~\ref{fig:gt_fig_base_ce} compares the base feature values and mean CE values for the representative case (N=25 Pareto solutions). The changes from base to CE are modest across all features. \textit{Nabe} and \textit{ramen} show the largest increases ($\approx+2$), \textit{sushi} shows a marginal increase ($\approx+1$), \textit{washoku} remains virtually unchanged, and \textit{izakaya} shows a marginal decrease ($\approx-1$).}

\subsubsection{Experiment 3.2: Educational Intervention Data}\label{subsec6b}

The proposed method is applied to real-world data, where the true model is unknown. This allowed us to investigate the impacts of specific variables, evaluate their {applicability}, and examine their potential applications in practical scenarios. {Academic achievement is shaped by a complex combination of interventions, with likely nonlinear thresholds and interaction effects (e.g., the benefit of individual educational support may depend on whether a student already records study time). Black-box models such as RF and MLP can capture these dependencies that a linear model cannot, motivating the need for CE methods to interpret their recommendations.}

Following is an overview of the data. The purpose of this survey was to investigate the factors influencing academic achievement among Japanese high school students. The survey was conducted in February 2022. The survey was conducted on 500 subjects (males and females aged 15 to 18) throughout Japan. The survey method was an Internet survey, and the sample was collected so that the sex and age ratios matched those of the national census. Academic achievement was measured on a 14-level self-reported ordinal scale corresponding to ranges of the deviation value (a standardized score commonly used in Japanese educational assessments); each level was converted to the midpoint of its deviation-value range (e.g., level~6 $\to$ 47.5) to yield a continuous target variable $\mathcal{Y}$. The remaining items---sex, age, and 19 intervention types ($T_1$, \dots, $T_{19}$)---were used as features.

{The 19 interventions $T_1$--$T_{19}$ are originally binary (0: not experienced, 1: experienced). For the optimization, these are relaxed to continuous values in $[0, 1]$, bounded by the inequality constraints $g_j$, so the optimizer never produces values outside $[0, 1]$. The optimized continuous values are retained as-is. Fractional values obtained by continuous relaxation may be interpreted as indicating the strength of interventions at the individual level or probabilistic interventions at the population level. Age and Sex are treated as fixed (equality constraint; no change is permitted).}

Descriptive statistics of the dataset are provided in Table~\ref{tab:descriptive_statistics} (Appendix~\ref{app:desc}).

The four models listed in Section~\ref{subsec5} are used. Based on this setup, we describe the results of our analysis. First, the accuracy of the model was verified (Table~\ref{tab:accuracy_models}). {All five models (including Stacking) had overlapping confidence intervals, satisfying the model-multiplicity condition. Among the four base models, the top-3 by MSE---Models 1, 4, and 2 (166.1, 168.6, and 180.8, respectively)---were selected for Method~3. Model~3 (LightGBM) was excluded as its MSE is nearly identical to Model~2, and including both would add redundancy without broadening model diversity. The Stacking model was reserved for Method~2.}

\begin{table}[htbp]
    \centering
    \caption{Accuracy of each model {on real data (mean $\pm$ SD over 20 splits)}}
    \label{tab:accuracy_models}
    \begin{tabular}{lc}
        \hline
        \textbf{Model} & \textbf{MSE (mean $\pm$ SD)} \\
        \hline
        Model 1 (LR)   & $166.139 \pm 16.083$ \\
        Model 2 (RF)   & $180.849 \pm 21.985$ \\
        Model 3 (LGB)  & $180.863 \pm 22.280$ \\
        Model 4 (MLP)  & $168.626 \pm 16.607$ \\
        Stacking Model & $167.107 \pm 17.667$ \\
        \hline
    \end{tabular}
\end{table}

Next, we evaluated these methods (Table~\ref{tab:evaluation index of ce}). Specifically, we set $y_t = \infty$. For {40} randomly selected base cases {(CE generation loop: seed~0; NSGA-II: seed~1)}, 10 CEs ($=S$) were generated for each case. Using these multiple CEs, the evaluation metrics described in Section~\ref{subsec4} were calculated. Finally, the mean values of these metrics across {40} cases were compared. The parameters were set to $C = 5$ and $\lambda = 5$ (the latter for Methods~1 and~2 only; Method~3 does not use $\lambda$). {The value $C = 5$ was chosen to allow up to 5 binary features to be changed simultaneously, which is a meaningful but not excessive degree of change given that there are 19 binary interventions. The value $\lambda = 5$ was set to balance validity and proximity for the educational intervention dataset (Methods~1 and~2 only).} For Method~3, NSGA-II was used with the same hyperparameters as Experiment~3.1. {The top-3 models by MSE---Models~1 (LR), 4 (MLP), and 2 (RF), with MSEs of 166.1, 168.6, and 180.8 respectively---were selected as the three objectives. Model~3 (LGB) was excluded due to near-identical MSE to Model~2.} In addition, we set the condition that the improvement is negative when the value is 1 and positive when the value is 0 in $T_1$--$T_{19}$. The flexibility to incorporate such conditions is one of the features of this method.

\begin{table*}[htbp]
    \centering
    \caption{{Evaluation metrics for CEs on real data (mean over 40 base cases, $S=10$ CEs each). Bold indicates the top-3 values per metric (all tied values included): highest for \textit{Val}, $\Delta$\textit{Val}, \textit{AV/D}, and \textit{AV/P}; lowest for \textit{Dissim} and \textit{Plaus}. TIR is not computable for real data (true model unknown). See Table~\ref{tab:ta2n1} for abbreviations.}}
    \label{tab:evaluation index of ce}
    \begin{tabular}{llrrrrrr}
        \toprule
        \textbf{Method} & & \textit{Val} & $\Delta$\textit{Val} & \textit{Dissim} & \textit{Plaus} & \textit{AV/D} & \textit{AV/P} \\
        \midrule
        M1 (LR)    & & \textbf{70.163} & $\mathbf{+20.663}$ &  3.004 &  4.232 & \textbf{23.353} & 16.580 \\
        M1 (RF)    & & 59.274 & $+9.774$ &  \textbf{1.839} &  \textbf{2.644} & \textbf{32.231} & \textbf{22.419} \\
        M1 (LGB)   & & \textbf{63.737} & $\mathbf{+14.237}$ &  \textbf{0.683} &  \textbf{0.799} & \textbf{93.284} & \textbf{79.723} \\
        M1 (MLP)   & & \textbf{63.578} & $\mathbf{+14.078}$ &  3.143 &  \textbf{1.967} & 20.230 & \textbf{32.327} \\
        M2 (Stack) & & 60.586 & $+11.086$ &  2.807 &  3.916 & 21.585 & 15.470 \\
        \multirow{3}{*}{M3 ($m$=3)} & (1) & 59.364 & $+9.864$ & \multirow{3}{*}{\textbf{2.646}} & \multirow{3}{*}{3.430} & \multirow{3}{*}{22.597} & \multirow{3}{*}{17.432} \\
         & (2) & 61.745 & $+12.245$ & & & & \\
         & (3) & 58.282 & $+8.782$ & & & & \\
        \bottomrule
    \end{tabular}
\end{table*}

{For \textit{Val} improvement, the single-model results (Method~1) varied substantially by the target model: Model~1 (Linear Regression) yielded the highest improvement ($\Delta\text{Val}=20.66$), followed by Model~3 (LightGBM; $\Delta\text{Val}=14.24$), Model~4 (MLP; $\Delta\text{Val}=14.08$), Model~2 (Random Forest; $\Delta\text{Val}=9.77$). Method~2 (stacking ensemble) produced $\Delta\text{Val}=11.09$. Method~3 (three-model MOO) achieved $\Delta\text{Val}$ of $+9.864$, $+12.245$, and $+8.782$ for Models~1 (LR), 4 (MLP), and 2 (RF), respectively, while explicitly optimizing across all three models simultaneously. Crucially, all three per-model $\Delta\textit{Val}$ values are positive, which is direct evidence of robustness under model multiplicity: the CEs simultaneously improve predicted outcomes across all three selected models. Method~1 provides no such guarantee for models other than its single target. For \textit{Dissim}, Method~1 targeting Model~3 (LGB) had the smallest value (0.68), with Method~1 (RF) achieving the second smallest (1.84); Method~3 achieved $\text{Dissim}=2.65$, the third smallest overall. For \textit{Plaus}, Method~3 ($\text{Plaus}=3.43$) is comparable to Method~1 (LR; 4.23). In summary, Method~3 achieves positive $\Delta\textit{Val}$ across all selected models without extreme deterioration in \textit{Dissim} or \textit{Plaus}, demonstrating its primary advantage of generating CEs that simultaneously satisfy multiple models.}

{The higher \textit{Dissim} of Method~3 relative to the most proximity-efficient single-model baselines (e.g., M1~(LGB): Dissim~=~0.68) is an expected consequence of multi-model robustness: CEs must simultaneously satisfy multiple prediction surfaces, which restricts the feasible region and naturally increases the required change from the base instance. This is the intended trade-off of the proposed method---model-agnostic validity across the multiplicity set takes precedence over minimizing distance to a single model.}

\begin{figure}[htbp]
    \centering
    \includegraphics[width=0.8\textwidth]{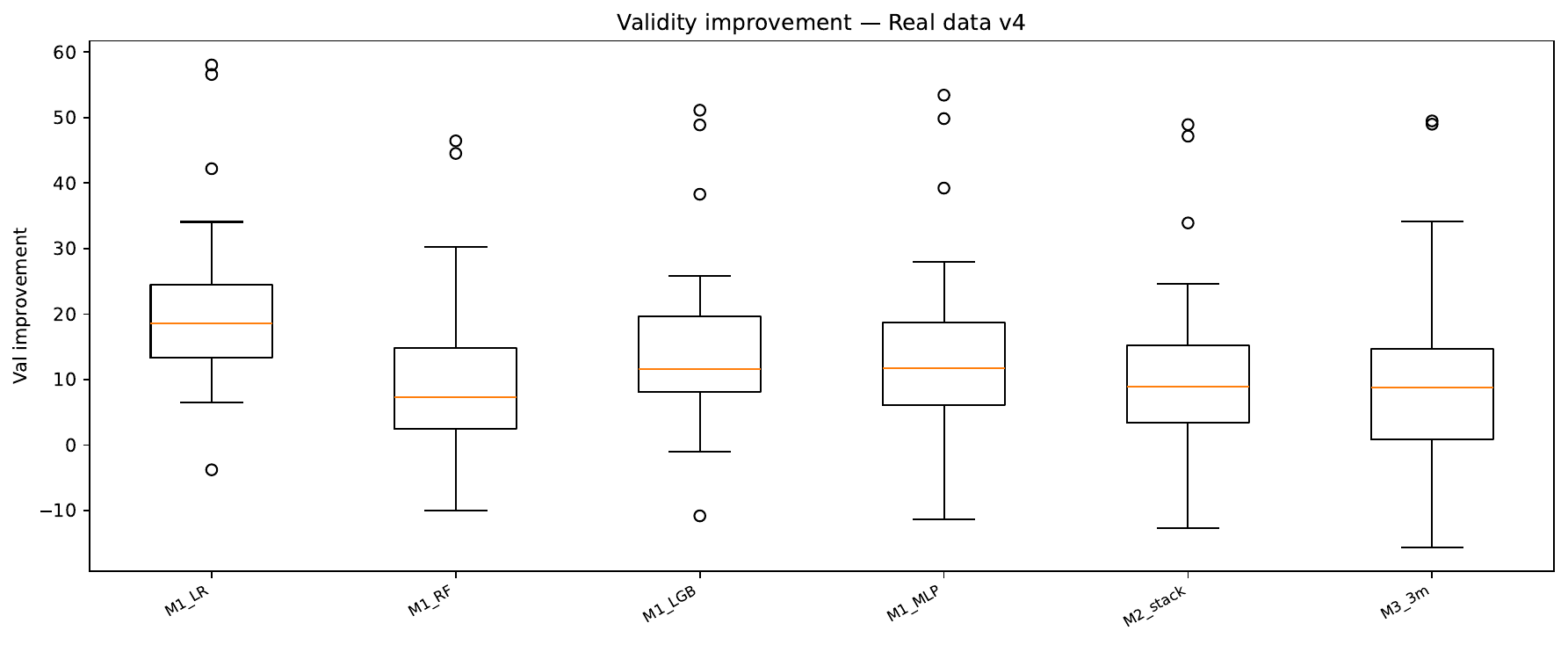}
    \caption{{Boxplot of Val improvement ($\Delta\text{Val}$) for each method on real data (40 base cases). Method labels correspond to: M1\_LR, M1\_RF, M1\_LGB, M1\_MLP (Method~1 targeting each single model), M2\_stack (Method~2), M3\_3m (Method~3, three-model MOO). Higher values indicate greater improvement over the base data.}}
    \label{fig:real_boxplot_val_improv}
\end{figure}

\begin{figure}[htbp]
    \centering
    \includegraphics[width=0.8\textwidth]{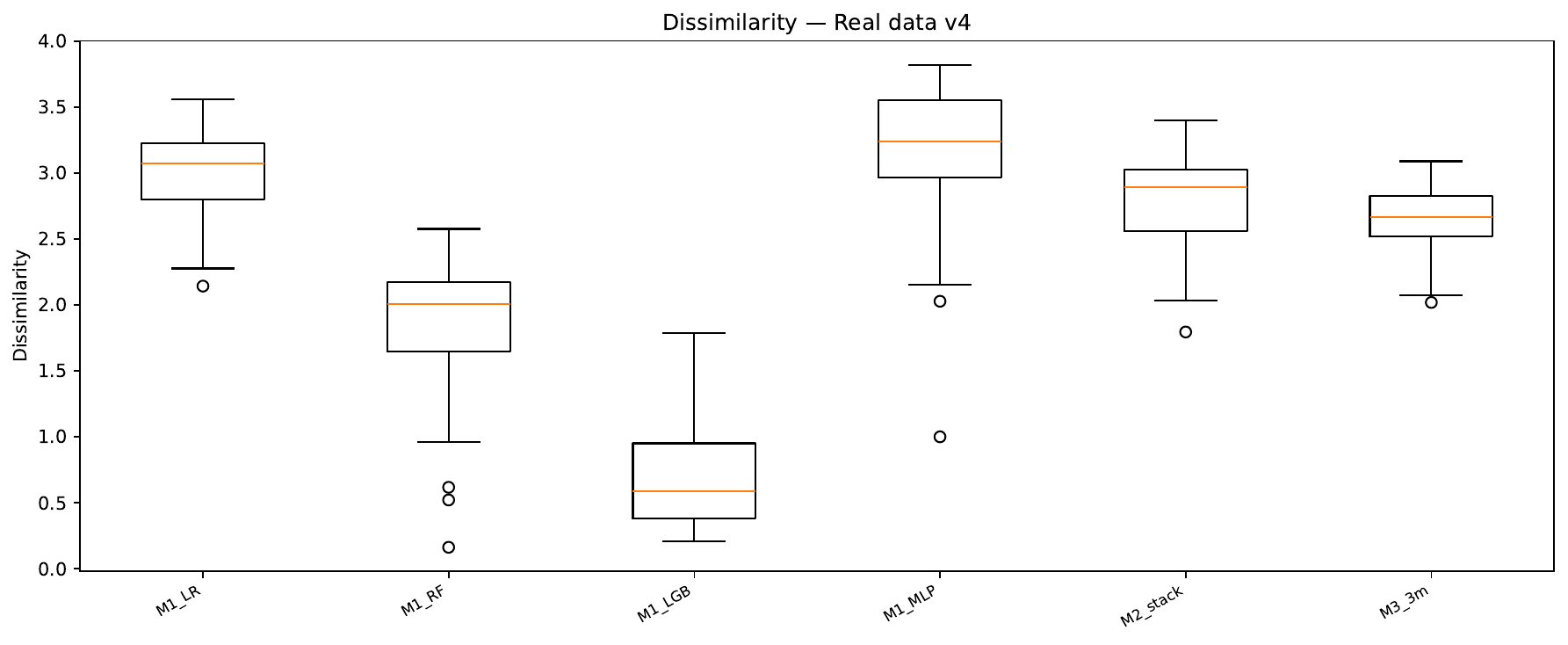}
    \caption{{Boxplot of Dissim for each method on real data (40 base cases). Lower values indicate that the CE requires fewer or smaller changes from the base data.}}
    \label{fig:real_boxplot_dissim}
\end{figure}

{Figure~\ref{fig:real_boxplot_val_improv} and Figure~\ref{fig:real_boxplot_dissim} show the distribution of Val improvement and Dissim across the 40 base cases for each method. In terms of Val improvement, M1\_LR achieves the highest median ($\approx+20$), followed by M1\_LGB and M1\_MLP ($\approx+14$). M3\_3m and M2\_stack achieve comparable medians ($\approx+10$--$11$) with moderate variance. In terms of Dissim, M1\_LGB shows the lowest Dissim (median $\approx0.7$), M1\_RF the second lowest ($\approx1.8$), and M3\_3m the third lowest ($\approx2.6$), reflecting the constraint of simultaneously satisfying multiple model objectives.}

\begin{figure}[htbp]
    \centering
    \includegraphics[width=\textwidth]{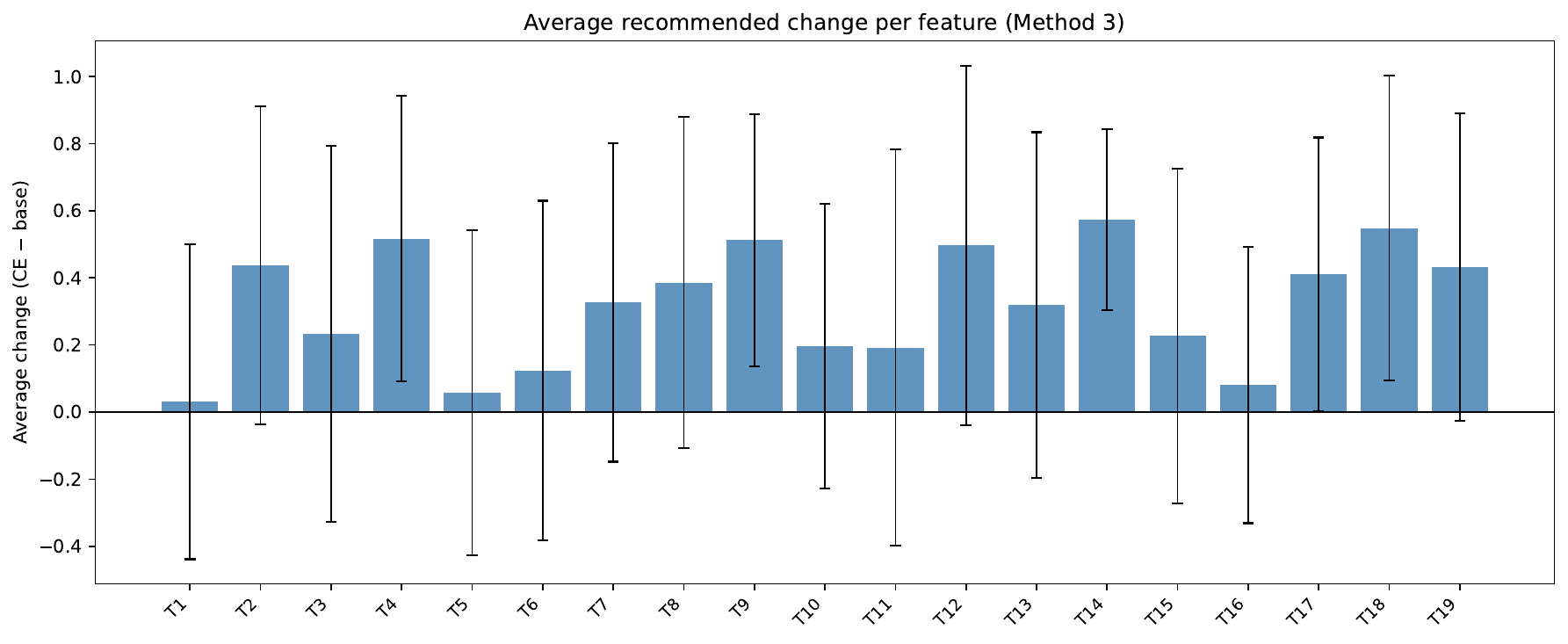}
    \captionsetup{justification=centering}
    \caption{{Average recommended change per feature (CE minus base value), aggregated as the mean of per-case means over 40 base cases. Positive values (blue) indicate that an increase is recommended; negative values (red) indicate a decrease; error bars indicate standard deviation across the 40 cases.}\\
    (vertical axis: average difference; error bars: standard deviation)}
    \label{fig:avg_excluding_baseline}
\end{figure}

{Figure~\ref{fig:avg_excluding_baseline} shows the average recommended change per feature (CE minus base value) across 40 base cases. All interventions receive positive recommendations, indicating that the CE framework consistently advises increasing exposure to educational interventions. The largest recommended increases are observed for $T_{14}$ (individual educational support), $T_{18}$ (supplementary classes outside of school), and $T_4$ (quantified science/arts wage gap). $T_1$ (college/grad wage premium knowledge), $T_5$ (employment and advancement rate information), and $T_{16}$ (workplace tours and internships) receive the smallest recommendations.}

Thus, it is possible to propose multiple improvement options for individual base data and identify variables whose effects are more robust based on their average values.

Taken together, the results across both simulation and real-data settings confirm the effectiveness and applicability of the proposed method, and motivate the following discussion of its broader implications.

\section{Discussion}\label{sec4}

In this study, we investigated robust CEs for machine learning, particularly for the problem of model multiplicity, which has become an issue in recent years. Specifically, we introduced the concept of Pareto improvement for robust CEs against model multiplicity and proposed the extraction of robust CEs using MOO. In addition, we proposed a robustness index.

The central finding is that multi-objective optimisation across multiple models is strictly necessary for robust CE generation under model multiplicity---proximity-first, single-model optimisation is insufficient. {The benchmark comparison in Experiment~2 (Section~\ref{subsec7}) confirms that the proposed method achieves the highest TIR in both simulation cases, outperforming Wachter-style (post-processing clipped) and DiCE baselines. In Case~2, Wachter achieves the smallest Dissim (2.609) and Plaus (6.623), but its TIR (0.635) is substantially lower than the proposed method's TIR (0.960). In Case~1, the proposed method achieves smaller Dissim (2.904 vs.\ 3.000) and Plaus (7.799 vs.\ 8.434) than Wachter while matching its TIR~=~1.000. These results indicate that proximity-first single-model optimisation does not improve multi-model robustness; the proposed method maintains competitive proximity while achieving substantially higher TIR, confirming that multi-objective optimisation across multiple models is the critical factor for robustness.} Extending to real-data settings where TIR cannot be computed directly, the real-data experiments confirmed \textit{Val} improvement across both datasets: the web search trend experiment (continuous features) demonstrated multi-model robust CE generation under bounded continuous variables, while the educational intervention experiment (binary features) identified key variables for improving academic achievement and confirmed diverse, actionable CE generation under model multiplicity. For a detailed comparison with related methods, including Pawelczyk et al.\ \cite{pawelczyk2020}, Leofante et al.\ \cite{leofante2023}, and Jiang et al.\ \cite{jiang2024recourse}, see Section~\ref{sec_related}.

The results above also suggest several methodological extensions worth pursuing. On the problem formulation side, when proximity is a primary concern, the distance constraint in the proposed method can be replaced by a weighted-sum loss term to enable explicit distance minimisation \cite{karimi2022}. On the solution-selection side, the Pareto improvement criterion naturally yields diverse trade-off solutions, allowing decision-makers to weight objectives according to their preferences and to visualise objective trade-offs. {Future work could further explore alternative selection criteria: under \textit{utilitarianism} \cite{harsanyi1977}, the CE maximizing the sum of improvements across all models would be preferred; under a \textit{Rawlsian} maximin criterion \cite{rawls1971}, the CE maximizing the worst-case improvement would be selected. These ethical frameworks may lead to different CE recommendations and their relative merits deserve further study.}

Beyond these extensions, future work will include determining which MOO algorithm is most suitable~\cite{gunantara2018} and investigating how results change when the optimization method is varied for \textit{Method 1} and \textit{Method 2}. The effects of the obtained CEs could also be evaluated through actual intervention experiments. Another potential avenue is to incorporate causality to enhance robustness~\cite{guidotti2024}, as causality pertains to the invariant structure of the underlying data. {Future work should further include reporting sparsity (L0 norm) as an additional evaluation metric, as suggested by Pawelczyk et al.\ \cite{pawelczyk2020}, and evaluation on publicly available benchmark datasets.} We believe this research can serve as a valuable foundation for various fields, including explainability in machine learning, decision-making, and action planning based on machine learning.

{
\section{Limitations}\label{sec5}

Several limitations of the current study should be acknowledged. First, TIR requires knowledge of the true function and can therefore only be computed on simulated data. On real data, we used $\Delta$\textit{Val} (improvement in predicted value) as a proxy; however, $\Delta$\textit{Val} measures only the model's predicted improvement, not the true outcome improvement, and may not reflect real-world effectiveness. A more principled real-world evaluation metric---such as one based on actual intervention outcomes---remains an important direction for future work. Second, the computational cost of NSGA-II grows with the number of features and models; scalability to high-dimensional datasets remains to be evaluated. Third, following Slack et al.\ \cite{slack2021}, we note that a diverse Pareto front of CEs could be misused by a malicious actor who cherry-picks a CE that appears favorable (fair-washing); mechanisms to mitigate this risk should be explored. Fourth, the continuous relaxation of binary intervention variables yields fractional CE values, interpreted here as intervention intensities or probabilistic interventions. In settings where interventions are strictly binary in practice, this interpretation may be less actionable; extending the framework with a post-processing rounding step or a mixed-integer formulation remains a direction for future work. Fifth, the sensitivity analysis of NSGA-II hyperparameters (population size and number of generations) was conducted only for simulation data (Table~\ref{tab:sensitivity_nsga2}); the corresponding analysis for real-data experiments remains a direction for future work.
}

{
\section{Related Work}\label{sec_related}

Counterfactual explanations (CEs) provide actionable recourse by recommending the minimal feature changes needed to alter a model's prediction \cite{karimi2022, guidotti2024, verma2024}. Beyond proximity, desiderata such as sparsity, plausibility, actionability, and diversity have been proposed \cite{guidotti2024}. A central open challenge is ensuring that CEs remain valid when conditions change---an issue known as robustness \cite{verma2024}.

Jiang et al.\ \cite{jiang2024} classified CE robustness into four categories: robustness against model changes (MC), model multiplicity (MM), noisy executions (NE), and input changes (IC), as outlined in Section~\ref{sec1}. MC has received the most attention: Pawelczyk et al.\ \cite{pawelczyk2022} showed that data-support methods better withstand retraining, Pawelczyk et al.\ \cite{pawelczyk2023} addressed robustness to noisy intervention execution, and Slack et al.\ \cite{slack2021} demonstrated that hill-climbing CE methods are vulnerable to adversarial input perturbations (fair-washing), where certain subgroups can receive up to 20$\times$ lower cost recourse under a slight perturbation. By contrast, robustness under MM---where multiple models achieve comparable accuracy on the same data---has been studied far less, yet poses a distinct practical challenge: a user's recommended recourse may be invalidated simply because a different model was deployed.

Pawelczyk et al.\ \cite{pawelczyk2020} provided the first formal treatment of CE robustness under MM, deriving cost bounds and showing that data-support methods are more robust to model multiplicity than sparse methods, albeit at higher per-model cost; however, their work is primarily analytical and proposes no generation algorithm. Leofante et al.\ \cite{leofante2023} proposed an exact mixed-integer linear programming (MILP)-based method to compute CEs valid across all models in a multiplicity set, but restricted it to homogeneous feedforward neural network classifiers with white-box access, and proved that the problem is NP-complete in general. Jiang et al.\ \cite{jiang2024recourse} proposed argumentative ensembling---using a Bipolar Argumentation Framework to select a consistent subset of pre-computed single-model CEs while accommodating user preferences---satisfying counterfactual validity and coherence, but not majority vote, and limited to classification.

The present work addresses the gaps left by the above. \textbf{(i)~Model-agnosticism:} unlike Leofante et al., our method requires no access to model internals and is not restricted to a specific architecture, making it applicable to any black-box model. \textbf{(ii)~Direct multi-model optimization:} unlike Jiang et al., CEs are generated \textit{directly} by simultaneous MOO across all selected models rather than by aggregating pre-computed single-model CEs; this avoids dependence on individual CE quality and enables a globally valid Pareto front. \textbf{(iii)~Regression and flexible constraints:} unlike all prior MM work, the framework handles continuous target variables and arbitrary inequality/equality constraints. Taken together, these properties fill a gap that existing methods leave open: model-agnostic, constraint-aware robust CE generation for MM in both classification and regression settings.

{Dandl et al.\ \cite{dandl2020} applied MOO to balance these competing criteria within a \textit{single} model, using NSGA-II \cite{deb2002}---a fast elitist non-dominated sorting genetic algorithm \cite{fliege2000, deb2002} that maintains a well-spread Pareto front through crowding distance. However, their work optimises within a single model and addresses neither model multiplicity nor CE robustness. The present work is, to our knowledge, the first to apply NSGA-II to the model-multiplicity CE problem, extending MOO from single-model desiderata balancing to cross-model robustness.}

{MOO algorithms differ substantially in how they handle multiple objectives. Classical scalarization approaches---weighted-sum and $\varepsilon$-constraint methods---yield a single solution per run \cite{deb2011} and require repeated runs with different parameter settings to approximate the Pareto front. Gradient-based methods \cite{mercier2018} converge quickly and require differentiable objectives, but may require additional mechanisms to obtain diverse solutions. In CE generation, models may be treated as black-box or involve non-differentiable components, making gradient-free evolutionary methods attractive in some settings. More recent evolutionary MOO developments include NSGA-III \cite{deb2014}, which uses structured reference points to handle many-objective problems more effectively, and Multi-Objective Evolutionary Algorithm based on Decomposition (MOEA/D) \cite{zhang2007}, which decomposes objectives into scalar subproblems solved collaboratively in a neighborhood structure. On the Bayesian side, the parallel noisy expected hypervolume improvement criterion (qNEHVI) \cite{daulton2021} has demonstrated strong performance for expensive multi-objective problems. NSGA-II remains appropriate for the present study because each objective evaluation is a single prediction call of a pre-trained model and is computationally inexpensive, reducing the primary advantage of surrogate-assisted methods in this setting, and the number of objectives ($m \leq 3$) is moderate, making the crowding-distance mechanism sufficient for maintaining diversity. Should the framework be extended to settings with expensive model evaluations or a larger number of objectives, qNEHVI could reduce the number of required evaluations and NSGA-III could improve Pareto front coverage.}
}

{
\section{Conclusion}

This paper proposed a multi-objective optimization framework for generating robust counterfactual explanations (CEs) under model multiplicity. By framing the CE generation problem as a Pareto improvement problem across multiple models, the proposed method ensures that CEs are simultaneously valid for all selected models, rather than optimizing for a single model or a weighted average. The key contributions are: (1) a principled definition of robustness under model multiplicity via Pareto optimality; (2) a model-agnostic framework supporting regression tasks and flexible constraints; (3) a diverse set of Pareto-optimal CEs enabling users to select solutions according to their preferences or constraints; and (4) application to two real-world datasets---web search trend data with continuous features and educational intervention data with binary features---demonstrating broad applicability.

Experiments on two simulated datasets confirmed that the proposed method (Method 3) consistently achieved higher TIR, comparable proximity, and superior plausibility compared to single-model (Method 1) and stacking-model (Method 2) baselines. {A benchmark comparison against Wachter-style (post-processing clipped) and DiCE baselines on simulation data further showed that the proposed method achieves the highest TIR in both simulation cases, confirming its advantage in multi-model robustness even when competing baselines achieve smaller Dissim and Plaus.} The real-data experiments confirmed the method's applicability across two domains: the web search trend experiment (continuous features) demonstrated robust CE generation under bounded continuous variables, while the educational intervention experiment identified key variables for improving academic achievement and showed that diverse, actionable CEs can be generated even for datasets with predominantly binary features.

We hope this work serves as a foundation for robust, fair, and actionable algorithmic recourse in safety-critical applications.
}

\begin{appendices}

\section{Descriptive Statistics of Educational Intervention Dataset}\label{app:desc}

\begin{table*}[htbp]
    \centering
    \caption{Descriptive statistics of the real dataset ($n=500$). For binary variables ($T_1$--$T_{19}$), the proportion of participants who experienced each intervention (\%) is reported. For continuous variables, mean $\pm$ SD is shown.}
    \label{tab:descriptive_statistics}
    \begin{tabular}{llp{7cm}r}
        \toprule
        \textbf{Variable} & \textbf{Type} & \textbf{Description} & \textbf{Value} \\
        \midrule
        Sex   & binary     & Male = 1, Female = 2 & 50.0\% (male) \\
        Age   & continuous & Integer (15--18)      & $16.9 \pm 0.9$ \\
        \midrule
        $T_1$  & binary & Knowledge1: college/grad wage premium over high school & 71.6\% \\
        $T_2$  & binary & Knowledge2: quantified lifetime wage difference (approx.\ 60--70M yen) & 24.2\% \\
        $T_3$  & binary & Knowledge3: science-track wages exceed arts-track on average & 40.2\% \\
        $T_4$  & binary & Knowledge4: quantified science/arts annual wage gap (approx.\ 400--600K yen) & 15.0\% \\
        $T_5$  & binary & Knowledge5: employment and advancement rates at postsecondary institutions & 30.6\% \\
        $T_6$  & binary & Considering the purpose of studying & 35.4\% \\
        $T_7$  & binary & Making and executing a study schedule & 32.2\% \\
        $T_8$  & binary & Recording study time & 18.0\% \\
        $T_9$  & binary & Recording study content & 14.8\% \\
        $T_{10}$ & binary & Sharing study content with friends & 12.8\% \\
        $T_{11}$ & binary & Group study and group work & 48.2\% \\
        $T_{12}$ & binary & Group leadership experience & 25.8\% \\
        $T_{13}$ & binary & Opinion presentation and presentation experience & 37.4\% \\
        $T_{14}$ & binary & Individual educational support & 9.8\% \\
        $T_{15}$ & binary & Career education & 19.6\% \\
        $T_{16}$ & binary & Workplace tours, internships, and related experiences & 20.6\% \\
        $T_{17}$ & binary & Introduction to seniors and other role models & 9.8\% \\
        $T_{18}$ & binary & Supplementary classes outside of school & 18.8\% \\
        $T_{19}$ & binary & Use of online educational materials & 29.2\% \\
        \bottomrule
    \end{tabular}
\end{table*}

\clearpage
\end{appendices}

\backmatter

\section*{Data Availability}
The data that support the findings of this study are available from the corresponding author upon reasonable request.

\bibliography{sn-bibliography}

\end{document}